\pdfoutput=1

\documentclass[11pt]{article}

\usepackage[]{arr_acl}

\usepackage{times}
\usepackage{latexsym}

\usepackage[T1]{fontenc}

\usepackage[utf8]{inputenc}

\usepackage{microtype}

\usepackage{inconsolata}

\usepackage{dirtytalk}

\usepackage{graphicx}
\usepackage{booktabs}
\usepackage{multirow}
\usepackage{tabularx}
\usepackage{makecell}
\usepackage{diagbox}
\usepackage[position=top]{subfig}
\usepackage{arydshln}
\usepackage{amssymb}
\usepackage[normalem]{ulem}  

\newcommand\redsout{\bgroup\markoverwith{\textcolor{red}{\rule[0.5ex]{2pt}{0.4pt}}}\ULon}

\newcommand{\taxtypeinfont}[1]{\textsc{#1}}
\newcommand{\taxtype}{\taxtypeinfont{Type}}
\newcommand{\taxtypes}{\taxtypeinfont{Type}s}

\title{Information Types in Product Reviews}

\author{Ori Shapira \\
  OriginAI \\
  Ramat Gan, Israel \\
  \texttt{obspp18@gmail.com} \\\And
  Yuval Pinter \\
  Ben-Gurion University of the Negev \\
  Beer Sheva, Israel \\
  \texttt{uvp@cs.bgu.ac.il}}

\begin{document}
\maketitle
\begin{abstract}
Information in text is communicated in a way that supports a goal for its reader.
Product reviews, for example, contain opinions, tips, product descriptions, and many other types of information that provide both direct insights, as well as unexpected signals for downstream applications.
We devise a typology of 24 communicative goals in sentences from the product review domain, and employ a zero-shot multi-label classifier that facilitates large-scale analyses of review data.
In our experiments, we find that the combination of classes in the typology forecasts helpfulness and sentiment of reviews, while supplying explanations for these decisions.
In addition, our typology enables analysis of review intent, effectiveness and rhetorical structure.
Characterizing the types of information in reviews unlocks many opportunities for more effective consumption of this genre.\footnote{Code and resources: \url{https://github.com/OriShapira/InfoTypes}}

\end{abstract}

\section{Introduction}
\label{sec_introduction}


A body of text is written with a purpose of conveying a message to its reader.
Information within the text is expressed in various formats and styles in order to fulfill this goal \citep{levy1979CommunicativeGoals}.
By identifying these methods of communication, a reader can focus on information of interest, or, conversely, obtain a clearer picture of the holistic intent of the full text.
In automatic applications, this is commonly accomplished by means of text classification that categorizes types of text spans.

Types of text differ across domains and communicative objectives.
For example, media coverage or debate statements attempt to convince of an ideological stance by using different frame typologies like `morality' or `legality'~\citep{naderi2017newsFrames, ali2022framingSurvey}.
Scientific articles contain snippets of texts purposed for explaining the discussed matter, like `background' or `results'~\citep{dernoncourt2017pubmed, cohan2019articleClassification}.
Product reviews often differ in sentiment to express an opinion regarding an aspect of the product~\citep{serranoguerrero2015sentAna, Yadav2020sentAna}.
Additionally, the length of a text influences its allowance for verbosity and detail~\citep{louis2014summaryContent}, and its rhetorical structure allows for a coherent arrangement of ideas within the text~\citep{mann1988rst}.
These characteristics, and others, provide means for classifying and analyzing text, at the levels of words, sentences, or documents.

In the e-commerce domain, product reviews assist a potential customer in learning about a product.
Reviewers share their experience by providing opinions and describing different aspects of the product.
They can suggest how to use the product, or compare it to alternative products. 
While doing so, they may vary their writing style to emphasize their viewpoint or argue for it.
We suggest that systematic identification of these facets can unlock novel explainable analyses of downstream applications.

\begin{figure}[t]
  \centering
  \includegraphics[width=0.9\linewidth]{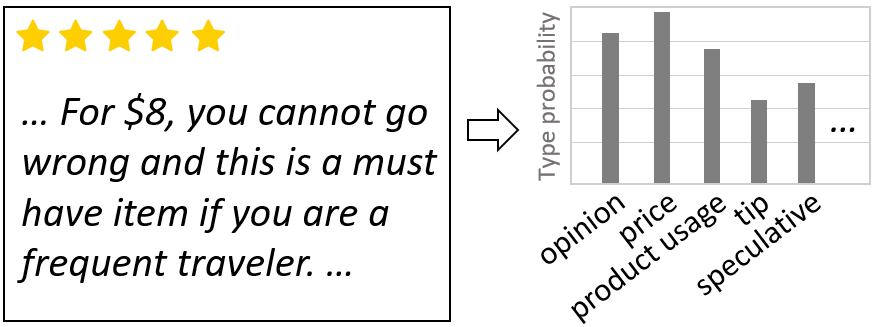}
  \caption{A sentence from a review, and the dominant information types communicated in it.}
  \label{fig_example}
\end{figure}

Content realization in product reviews has been extensively studied through, for example, aspect extraction and sentiment analysis~\citep{Tubishat2018aspExtr, xu2019bertSentAna}, but very little research has centered on \textit{types} of content within reviews, excluding `tips' and `arguments'~\citep{hirsch2021producttips, liu2017argsForHelpfullness}.
We introduce a \textit{broad} typology of sentence types for product reviews (\S{\ref{sec_taxonomy}}) consisting of 24 classes, including both functional types (e.g.,~`opinion', `product usage') as well as markers of linguistic style (e.g.,~`sarcasm', `imagery').
\autoref{fig_example} presents an illustration of an example review sentence with some of its plausible types.
More examples are available in \autoref{tab_example_annotations} in the appendix.

We conduct large-scale experimentation and analysis of our typology on product reviews by leveraging zero-shot capabilities of a large language model, \texttt{flan-T5-xxl}~\citep{wei2022flan} (\S{\ref{sec_prediction}}).
We then demonstrate how sentence type labels alone, without further access to the text itself, provide a strong signal for major downstream tasks, including review helpfulness and sentiment analysis (\S{\ref{sec_experiments}}).
We show that revealing the types of information present in reviews and their summaries provides 
insight into the structure of subjective text and the nature of information readers find useful (\S{\ref{sec_analysis}}).
The enriched understanding of such texts can in turn lead to improved consumption and enhanced modeling of relevant applications.




\section{A Typology of Review Sentences}
\label{sec_taxonomy}
We aim to define a typology that classifies the vast information available in reviews into types that would, in turn, provide insights to review consumers like potential customers or product vendors~\citep{Rohrdantz2016review_insights}.
Establishing a new typology relies on intuition and theory of the domain, as well as empirical analysis of data \citep{Nickerson2013taxonomydevelop}, and requires an iterative process that combines these two fundamental ingredients. This approach is employed across diverse domains, such as conversational social cues \citep[e.g.,][]{Feine2019conversationTaxonomy} and psychological well-being \citep[e.g.,][]{Desmet2020psychoTypology}.

Following this common practice and based on \citet{Nickerson2013taxonomydevelop}, the starting point for defining our typology is a set of existing text types previously researched individually, such as tips~\citep{hirsch2021producttips}, opinions~\citep{vinodhini2012opinionmining}, toxicity~\citep{Djuric2015hatespeech}, sarcasm~\citep{aditya2017sarcasm}, and the distinction between product, delivery, and seller descriptions~\citep{Bhattacharya2020seller_delivery_classification}.
\citet{Rohrdantz2016review_insights} analyzed reviews and defined a theoretical taxonomy of review types that provides us with further inspiration. The next step requires examining data instances and iteratively deriving the information types in review sentences.

To this end, we sampled reviews from eight categories of the Amazon Product Review Dataset \citep{he2016reviewsDS}, and split them into sentences (technical details in Appendix~\ref{sec_appendix_taxonomy}).
The variability of product categories ensures representation of as many sentence types as possible.
One of the authors manually inspected a random sample of several hundred sentences and accumulated sentence types, always marking a sentence with at least one type (\textit{rhetorical} being a catch-all default tag).
The typology was iteratively fine-tuned during this tagging process, as some types were only noticed after several occurrences.
The process concluded as soon as no new types were encountered for over 100 sentences, yielding a total of roughly 550 reviewed sentences.
The resulting list of types was examined by all authors and slightly refined, resulting in the final typology of 24 \taxtypes{}: \vspace{4pt} \\ \noindent\fbox{%
    \parbox{\columnwidth - 2\fboxsep}{%
        \textit{opinion, opinion with reason, improvement desire, comparative, comparative general, buy decision, speculative, personal usage, situation, setup, tip, product usage, product description, price, compatibility, personal info, general info, comparative seller, seller experience, delivery experience, imagery, sarcasm, rhetorical, inappropriate}
    }%
} \vspace{0.5pt} \\ (see \autoref{tab_taxonomy_types} in the appendix for explanations and example sentences of the \taxtypes{}).
While not hierarchical, the \taxtypes{} can be grouped by mutual semantic characteristics into \textit{coarse-grained} types, such as being objective, subjective or stylistic in nature, with some groups partially overlapping (see \autoref{tab_taxonomy_groups}).
These groupings will come in handy in our analyses in Sections \ref{sec_prediction}, \ref{sec_experiments} and \ref{sec_analysis}.

\begin{table}[t]
\centering
    \resizebox{\columnwidth}{!}{
    \begin{tabular}{ll}
        \toprule
        \textbf{Group} & \textbf{\taxtypes{}} \\
        \midrule
        subjective     & \begin{tabular}[c]{@{}l@{}}opinion, opinion\_with\_reason,\\ improvement\_desire, buy\_decision,\\ speculative, seller\_experience, delivery\_experience\end{tabular} \\
        \cdashline{1-1}
        \cline{2-2}
        opinions       & opinion, opinion\_with\_reason \\
        \hline
        objective      & \begin{tabular}[c]{@{}l@{}}comparative, comparative\_general, personal\_usage,\\ situation, setup, tip, product\_usage, product\_description,\\ price, compatibility, general\_info, comparative\_seller\end{tabular} \\
        \cdashline{1-1}
        \cline{2-2}
        description    & \begin{tabular}[c]{@{}l@{}}setup, tip, product\_usage, product\_description,\\ price, compatibility\end{tabular} \\
        \cdashline{1-1}
        \cline{2-2}
        comparisons    & \begin{tabular}[c]{@{}l@{}}comparative, comparative\_general,\\ comparative\_seller \end{tabular} \\
        \hline
        personal       & personal\_usage, personal\_info \\
        \hline
        non\_product   & \begin{tabular}[c]{@{}l@{}}personal\_info, general\_info, comparative\_seller,\\ seller\_experience, delivery\_experience\end{tabular} \\
        \hline
        stylistic     & imagery, sarcasm, rhetorical, inappropriate \\
        \bottomrule
    \end{tabular}}
    \caption{The \taxtypes{} in our typology grouped by mutual semantic characteristics (into coarse-grained types). Some groups partly overlap or are fully contained in others (dashed sections).}
    \label{tab_taxonomy_groups}
\end{table}


While the typology was defined systematically and carefully, we note that there are certainly other exhaustive typologies that could be established for categorizing information types in reviews.
Our main goal is to reveal the many advantages of identifying content types in reviews using a highly variable typology, whether for understanding review style or for informing existing downstream classification tasks, as we will show in our analyses. A typology different from ours may yield slightly different outcomes, but the general concept and usability are expected to be persistent.

\section{Detecting Sentence Types}
\label{sec_prediction}
In order to facilitate analysis and experiment with reviews, and to be able to do so at a large scale, we implement a sentence-level multi-label classifier for our typology.
Traditionally, this would require collecting an annotated dataset that reasonably represents the sentence \taxtype{} distribution.
However, considering the large amount of classes, some of which are quite rare in the data, this would be a challenging undertaking. 
Fortunately, these days we have powerful tools at our disposal which can avoid this process. 
Recent advances in pretrained large language models (LLMs) have shown strong zero-shot capabilities~\citep{wei2022flan}, which we leverage for our multi-label classification task.

\subsection{Classification Model}
\label{sec_prediction_model}

\paragraph{Development data.}
We start by collecting a small seed set of annotated sentences on which we can perform trial runs on candidate LLMs, testing the adequacy of different prompts.
To that end, we sampled a set of sentences from Amazon reviews \citep{he2016reviewsDS}, separate from the ones used to define the typology (\S\ref{sec_taxonomy}). 
Following an initial attempt to crowdsource this process, which did not fare well,\footnote{Crowdsourcing produced sparse and noisy annotations. 
The task requires concentration and a certain level of expertise, that cannot be expected from standard crowdworkers.}
we resorted to an internal annotation procedure.
Two authors annotated 123 sentences using the full \taxtype{} set defined, with an initial inter-annotator reliability (Cohen’s $\kappa$) of 0.71.
Cases with no agreement (9 sentences) or partial \taxtype{} overlap (37) were resolved through discussion. Due to the strong agreement, one of the authors then proceeded to annotate 300 new sentences for the development set.

\paragraph{Model.}
Using the development set, we evaluated the prediction capabilities of several instruct-based LLMs: FLAN-T5~\citep[\texttt{flan-t5-\{xl, xxl\}},][]{chung2022flan}, UL2~\citep[\texttt{flan-ul2},][]{tay2023ul2}, and Jurassic~\citep[\texttt{j2-jumbo-instruct},][]{ai212023jurassic}, using two or three temperature settings for each.
Preliminary experiments clearly showed that a single-label classification setup for each sentence \taxtype{} produces much better results than a multi-label setup attempting to annotate all 24 \taxtypes{} at once.
In the single-label setup, a classification request is sent per \taxtype{}, requesting a \say{yes} or \say{no} answer.
Each model required slightly different prompt formats in order to output relevant replies. 
Some models (\texttt{flan-t5-xl} and \texttt{j2-jumbo-instruct}) somewhat improved when they were asked to explain their answer~\citep{wei2022chainofthought, liu2023prompt}.
We found that \texttt{flan-t5-xxl} and \texttt{flan-ul2} with temperature 0.3 provided the best responses, both in terms of output-format consistency and correctness on the development data, affirming FLAN's span-based-response training procedure optimized for classification tasks~\citep{wei2022flan}.
Of the two flavors, we opted for the faster \texttt{flan-t5-xxl} for the rest of our procedure.

We control for the variance in model responses by querying each prompt 10 times, and setting a final score for the respective \taxtype{} as the proportion of \say{yes} responses.\footnote{This protocol was followed as a workaround for not having access to LLM output probabilities. Querying 30 times produced similar performance on the development set. Results using this protocol were highly replicable when re-run.}
See Appendix \ref{sec_appendix_predictor} for model details and comparisons, and for the prompt templates used.




\subsection{Classification Model Quality}
\label{sec_prediction_quality}
We now measure the model's ability to predict all \taxtypes{} of the proposed typology in a multi-class multi-label setup.

\paragraph{Test data.}
We first need to collect test data on which to evaluate the model. Due to the long-tail effect, a small sample of sentences representing the actual data distribution would likely contain very few instances for certain \taxtypes{} (e.g., \textit{comparative seller} and \textit{inappropriate}).
We thus prepared a more \taxtype{}-balanced test set in the following manner: 
We ran the classifier model from \S{\ref{sec_prediction_model}} on a large sample of sentences from Amazon reviews~\citep{he2016reviewsDS},\footnote{The categories of the reviews are partly different from those in the development set.} as described in Appendix~\ref{sec_appendix_experiments_analysis}, and for each of the 24 \taxtypes{}, sampled 10 random sentences whose prediction scores are above the optimal threshold determined on the development set.
The original annotating author (\S{\ref{sec_prediction_model}}) then annotated these 240 sentences, \textit{without} access to the model's predicted labels, to create the test set.\footnote{While this two-step procedure indeed resulted in a higher count of under-represented types, the \taxtype{}-distribution Pearson correlation between the dev and test sets is still high (0.91), due to the multi-label nature of many sentences. See \autoref{tab_test_scores_per_type} in the appendix for the \taxtype{}-distributions in the development and test sets.}

\paragraph{Results.}
As commonly conducted in multi-class multi-label classification, we measure the accuracy of the classifier using a macro-$F_1$ score, i.e., the non-weighted average of \taxtype{}-level $F_1$ scores on the sentences in the test set. Our classifier achieves a macro-$F_1$ of 56.7, with per-\taxtype{} $F_1$ scores reported in \autoref{tab_test_scores_per_type} in the appendix. Compare this to a random baseline (that guesses with 50\% chance) that receives a macro-$F_1$ of 40.9.

It is apparent that some \taxtypes{} are predicted better than others; however on manual inspection, most of the incorrect predictions are borderline cases (see \autoref{tab_example_annotations} in the appendix for examples). This is further corroborated by computing a coarse-grained score (mapping the 24 \taxtypes{} to the 8 groups from \autoref{tab_taxonomy_groups}), which produces a macro-$F_1$ score of 79.7 (with an expected random baseline score of 47.4). I.e., \taxtypes{} of similar semantic characteristics may be wrongly predicted in reference to the gold label, but are still instead identified as closely related \taxtypes{}.

The results demonstrate the usability of the model to generally classify review sentences according to our typology. While future work could further improve the classifier, either with training data for fine-tuning, more prompt engineering, or better base models, our current model is reliable enough for our analyses presented next.\footnote{We note that some of our \taxtypes{} have been studied before (e.g., \textit{tips} and \textit{opinions}). In Appendix \ref{sec_appendix_predictor_eval_type_specific} we present results for classification of these individual labels, reinforcing the usefulness of our classifier, as a generalization of this previous work.}
\section{Downstream Review Classification}
\label{sec_experiments}
We turn to exemplify how the combination of the varying \taxtypes{} in the typology serves as a strong \say{feature set} for addressing many downstream tasks that apply to review data.
Here, we address the tasks of review helpfulness, sentence helpfulness, and sentiment analysis.
We build models that use only the predicted typology labels as input (without further access to the text itself), with the goal
not to outperform state of the art results, but rather to exhibit the strong signals concealed within the varying \taxtypes{} and their combinations.

\begin{table}[t]
    \centering
    \resizebox{\columnwidth}{!}{
    \begin{tabular}{clccc}
        \toprule
                 & & \multirow{2}{*}{\makecell{Helpful\\Reviews}} & \multirow{2}{*}{\makecell{Helpful\\Sentences}} & \multirow{2}{*}{\makecell{Sentiment\\Analysis}} \\
        \multicolumn{2}{l}{Method} & & \\
        \midrule
        \multirow{6}{*}{\rotatebox[origin=c]{90}{{\small\taxtype{} Set with SVM}}} & subjective      & 69.6 & 82.1 & 83.7 \\
        & ~~~(\textit{opinion} only)    & 48.7 & 72.7 & 80.2 \\
        & ~~~(\textit{op. w/rsn.} only) & 67.4 & 80.9 & 80.2 \\
        & objective       & 67.3 & 83.3 & 80.2 \\
        & stylistic       & 63.1 & 63.7 & 84.0 \\
        \cmidrule{2-5}
        & \textbf{All \taxtypes{}} & 72.6 & 88.3 & 88.1 \\
        \midrule
        \multicolumn{2}{l}{Coarse-grained types w/SVM} & 67.7 & 85.9 & 80.4 \\
        \multicolumn{2}{l}{Random (50-50)}             & 50.0 & 50.0 & 50.0 \\
        \multicolumn{2}{l}{Random (proportion known)}  & 50.0 & 51.0 & 68.3 \\
        \bottomrule
    \end{tabular}}
    \caption{Accuracy scores (\%) on the classification tasks using an SVM with different subsets of \taxtypes{} (\autoref{tab_taxonomy_groups}) from the typology (top). The combination of all \taxtypes{} best predicts helpfulness and sentiment. The bottom part shows prediction results using the coarse-grained types from \autoref{tab_taxonomy_groups} as features to an SVM, and the results of random predictors. Full results are in \autoref{tab_classification_results_full} in the appendix.}
    \label{tab_classification_results}
\end{table}

\subsection{Review Helpfulness}
\label{sec_experiments_review_helpfulness}

Predicting helpfulness of reviews is a popular task~\citep{ocampodiaz2018helpfulnessFeatures} due to its essential role in highlighting useful information to customers on e-commerce sites.
In order to assess how different subsets of the \taxtypes{} assist with helpfulness prediction, we extract Amazon reviews~\citep{he2016reviewsDS} and identify helpful and non-helpful reviews based on \say{up-votes} and \say{down-votes} on the reviews ($\sim$1000 balanced reviews, details in Appendix \ref{sec_appendix_experiments_review_helpfulness}).
Each sentence from a review
is input to our classifier (\S\ref{sec_prediction_model}), producing a 24-dimensional vector of \taxtype{} probabilities.
The average of these vectors then represents the review-level \taxtype{} probabilities.\footnote{Notice that each \taxtype{} has an independent probability, and probabilities in a vector do not add up to 100\%, since this is a multi-label classification setup.}
We reuse this averaging procedure throughout the rest of our experiments and analyses.

We then train a binary SVM classifier~\citep{Cortes1995svm} over the vectors to classify reviews as helpful or non-helpful, and evaluate its performance through cross-validation. We report classification accuracy, as is common in binary classification setups~\citep{Yadav2020sentAna}.
As seen in \autoref{tab_classification_results} (leftmost results column), using the full \taxtypes{} vector as an input produces a respectable accuracy of 72.6\%, while applying only certain subsets of \taxtypes{} results in significantly ($p<0.001$) lower scores, emphasizing the importance of the variability in the typology.

On further analysis, the results indicate that the \textit{opinion with reason} \taxtype{} is considerably more vital for helpfulness of a review than the \textit{opinion} \taxtype{}, possibly because providing reasoning for an opinion is more convincing.
This is noticeable through the immense gap between the second and third rows of \autoref{tab_classification_results}, and is further illustrated in Figure~\ref{fig_classification_vectors_hr}, which shows probability scores for each \taxtype{} aggregated at the review level and stratified based on review helpfulness.
Generally, the subjective \taxtypes{} (which include opinions---refer to \autoref{tab_taxonomy_groups}) most strongly signal helpfulness, however the combination of all \taxtypes{} performs best.
Results on all \taxtype{} subsets are available in \autoref{tab_classification_results_full} in the appendix.

The bottom section of \autoref{tab_classification_results} presents the results when using the coarse-grained types (from \autoref{tab_taxonomy_groups}) as features for an SVM model, where the predicted fine-grained \taxtypes{} are mapped to their respective coarse-grained types. The fine-grained typology yields better results than the coarse-grained one, reiterating its benefit. The two random binary baselines are for lower-bound reference. The first baseline randomly chooses an answer at 50\%, while the other chooses an answer assuming it knows the distribution of the gold labels.

\begin{table}[t]
    \centering
    \resizebox{\columnwidth}{!}{
    \begin{tabular}{clccc}
        \toprule
        \multicolumn{2}{l}{Method} & MSE $
        \downarrow$ & PC $\uparrow$ & N@1 $\uparrow$ \\
        \midrule
        \multirow{6}{*}{\rotatebox[origin=c]{90}{{\small \taxtype{} Set w/ Log. Regr.}}}
        & subjective      & 0.096 & 0.66 & 0.86  \\
        & ~~~(\textit{opinion} only)    & 0.121 & 0.53 & 0.66  \\
        & ~~~(\textit{op. w/rsn.} only) & 0.116 & 0.56 & 0.74  \\
        & objective       & 0.103 & 0.63 & 0.86  \\
        & stylistic      & 0.149 & 0.33 & 0.57  \\
        \cmidrule{2-5}
        & All \taxtypes{} & 0.075 & 0.75 & 0.86  \\
        \midrule
        \multirow{3}{*}{\rotatebox[origin=c]{90}{{\tiny\citeauthor{gamzu2021helpfulsentences}}}} & Baseline 1 {\small(TF-IDF)}        & 0.090 & 0.63 & 0.91 \\
        & Baseline 2 {\small(ST-RIDGE)}        & 0.062 & 0.78 & 0.94 \\
        & Best {\small(BERT)}            & 0.053 & 0.84 & 0.95 \\
        \bottomrule
    \end{tabular}}
    \caption{Mean squared error, Pearson correlation and NDCG@1 scores on the helpful sentence scoring task using logistic regression with different subsets of \taxtypes{} from the typology (\autoref{tab_taxonomy_groups}), compared to results of \citet{gamzu2021helpfulsentences}.
    Full results are in \autoref{tab_helpful_sentences_results_full} in the appendix.}
    \label{tab_helpful_sentences_results}
\end{table}

\subsection{Review Sentence Helpfulness}
\label{sec_experiments_sentence_helpfulness}

The next experiment assesses helpfulness of individual sentences in reviews (as opposed to full reviews), a task introduced by \citet{gamzu2021helpfulsentences}
as a form of extreme product review summarization.
Their review sentence dataset is annotated for helpfulness along a range from 0 (not helpful) to 2 (very helpful).
We produce the \taxtype{} vectors for the sentences and train a linear regression model to predict helpful sentence scores.
In addition, we generate a binary classification task by separating the data into subsets of helpful and unhelpful sentences, taken from the top and bottom tertiles (see Appendix \ref{sec_appendix_experiments_sentence_helpfulness}), on which we train an SVM model.

We compare our results for the regression task with those of \citet{gamzu2021helpfulsentences} in \autoref{tab_helpful_sentences_results}.
As in the review helpfulness task in \S{\ref{sec_experiments_review_helpfulness}}, it is apparent that the \taxtype{}-vectors provide a strong signal for predicting helpfulness also on the sentence level.
The results of the binary classification task, shown in \autoref{tab_classification_results} (center results column), reinforce this conclusion. 
When comparing the aggregate \taxtype{} probability vectors of helpful sentences to those of unhelpful sentences, Figure~\ref{fig_classification_vectors_hs} exhibits large contrast on the \textit{opinion with reason}, \textit{personal info} and \textit{product description} \taxtypes{}.
The \textit{subjective} \taxtype{} subset provides the strongest signal, however the combination of all fine-grained \taxtypes{} substantially improves over all subsets, again emphasizing the importance of the variability in our typology.

\subsection{Review Sentiment Polarity}
\label{sec_experiments_sentiment}

Over the years, the customer review domain has been central in the task of sentiment analysis \citep{chen2017sentAnaWithTypeClassification, Tesfagergish2022zeroshot_classification}. Understanding the sentiment in reviews assists in surfacing positive and negative criticism on products or services. This in turn aids new customers in making more informed purchasing decisions, and concurrently facilitates improvement on the vendors' side.

While, at first glance, the information \taxtypes{} in our typology should not associate with sentiment in any obvious manner, we perform a similar classification assessment as in the previous experiments.
In this case, we randomly extract 5,000 Amazon reviews~\citep{he2016reviewsDS} and group them into positive and negative reviews (review ratings of $\{4,5\}$, or $\{1,2,3\}$ respectively, as commonly handled~\citep{Shivaprasad2017SentAnaRev}).
Again training and evaluating an SVM through cross-validaition, we classify the reviews into their sentiment polarities (details in Appendix \ref{sec_appendix_experiments_sentiment}).

\begin{figure}
    \centering
    \subfloat[\textbf{Reviews} grouped by \textbf{helpfulness}.]{%
        \includegraphics[clip,width=.95\linewidth]{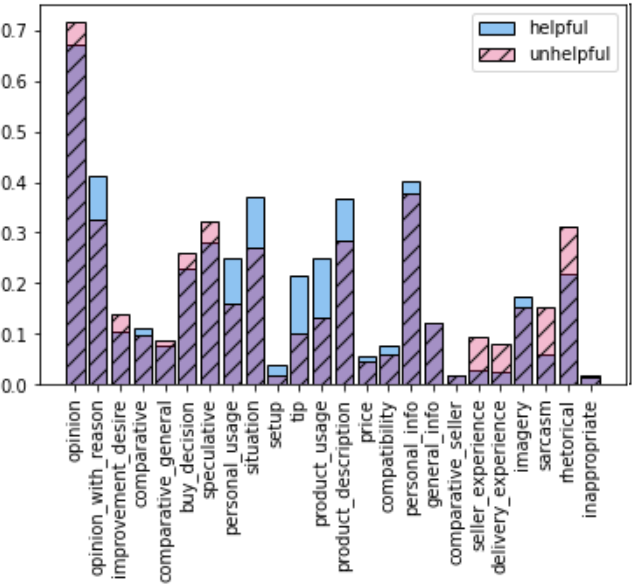}%
        \label{fig_classification_vectors_hr}
    }
    
    \subfloat[\textbf{Review sentences} grouped by \textbf{helpfulness}.]{%
        \includegraphics[clip,width=.95\linewidth]{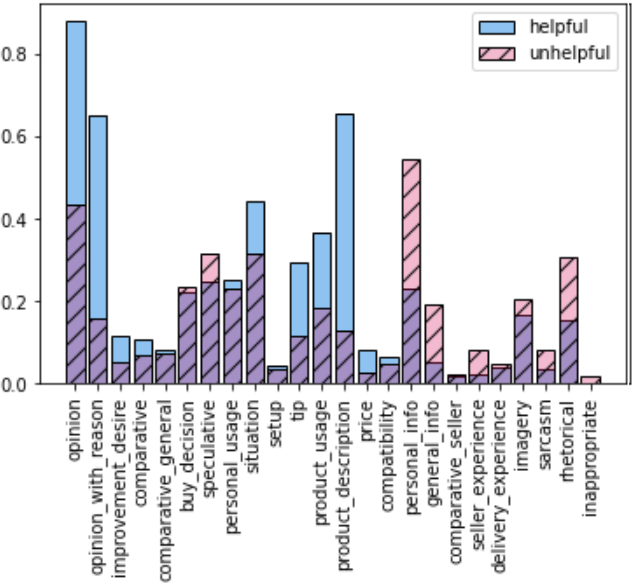}%
        \label{fig_classification_vectors_hs}
    }
    
    \subfloat[\textbf{Reviews} grouped by \textbf{sentiment}.]{%
        \includegraphics[clip,width=.95\linewidth]{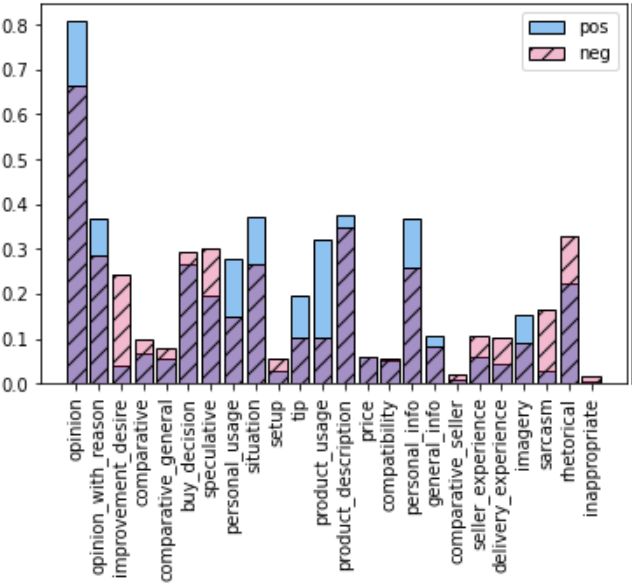}%
        \label{fig_classification_vectors_sentiment}
    }
    
    \caption{Aggregated \taxtype{} probability vectors over three classification tasks.
    Review-level probability vectors (\ref{fig_classification_vectors_hr} and \ref{fig_classification_vectors_sentiment}) are computed as the average of the sentence vectors contained in the review.}
    \label{fig_classification_vectors}
\end{figure}

Surprisingly, the SVM classifier is able to separate positive from negative reviews with an accuracy of 88.1\%, as seen in \autoref{tab_classification_results} (rightmost column).
Specifically, the combination of the subjective and stylistic \taxtype{}-sets provides the signal, while all other \taxtype{}-sets fail at identifying a distinction (the SVM classifier simply marks all sentences as positive). On closer inspection, Figure~\ref{fig_classification_vectors_sentiment} shows that the \textit{improvement desire} and stylistic \taxtypes{} indeed differ markedly between the positive and negative reviews, which offers a plausible explanation for the results.

While many models addressing the sentiment analysis task lack explainability for the sentimental decision~\citep{Adak2022sentimentExplainability}, the information \taxtypes{} give a hint at what induces the sentimental polarity, and can further direct attention at specific features. In this case, by focusing on sentences that express an improvement desire, review readers can more efficiently understand what to expect from the product, and can refrain from reading full negative reviews that are rich in rhetorical information.

The three experiments above suggest that the type of content in reviews offers strong indicators for helpfulness and sentiment, and that the residual textual content past the types may be only of secondary importance. Moreover, utilizing fine-grained types provides stronger signals than coarse-grained ones, since, apparently, \textit{specific} types of content distinctly influence humans' perception of helpfulness and sentiment.


%
%
%
%
%

\section{Analysis of Reviews and Summaries}
\label{sec_analysis}
As illustrated in \S{\ref{sec_experiments}}, our typology offers an instrument to analyze product-related texts and interpret differences between subsets of texts. In this section we focus on analyzing the structure of reviews and of review summaries. The analysis sheds light on the content patterns that reviewers choose to voice and what summarizers focus on when space is limited.

\begin{figure}
  \centering
  \includegraphics[width=0.9\linewidth]{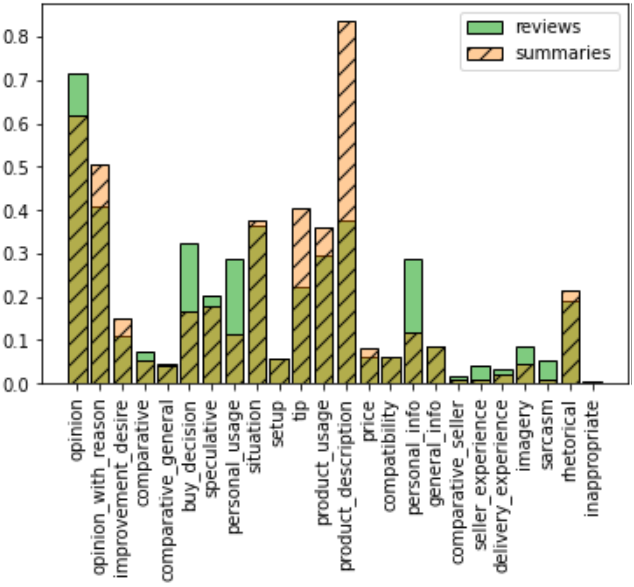}
  \caption{Aggregated \taxtype{} probability vectors in AmaSum~\citep{brazinskas2021amasum} reviews and summaries.}
  \label{fig_review_vs_summaries_vectors}
\end{figure}

\paragraph{Data.}
The AmaSum product reviews summarization dataset~\citep{brazinskas2021amasum} consists of about 31K products, each associated with a list of reviews and a reference summary taken from online platforms for professional product reviews.
Each summary comprises \emph{verdict}, \emph{pros} and \emph{cons} sections. 
We perform some filtering heuristics and then sample 100 products with their reviews and summaries, representing five different product categories (see App.~\ref{sec_appendix_experiments_analysis}), for a total of 7,729 reviews and 100 summaries.


\begin{table}[t]
    \centering
    \resizebox{\columnwidth}{!}{
    \begin{tabular}{cccc}
        \toprule
        Helpful-S & Unhelpful-S & Helpful-R & Unhelpful-R \\
        \midrule
        $0.922^{**}$ & $0.359^\dag$ & $0.795^{**}$ & $0.602^{**}$ \\
        \bottomrule
    \end{tabular}}
    \caption{Pearson correlation between \taxtype{} vectors of review sentences (S) or reviews (R), and summaries. 
    \\ $**$ $p < 0.001$, $\dag$ $p = 0.085$}
    \label{tab_summary_helpfulness_correlation}
\end{table}

\subsection{Reviews vs. Summaries}
We compare \taxtype{} prominence between reviews and summaries by aggregating their
probability vectors (averaging review- or summary-level \taxtype{} vectors),
presented in \autoref{fig_review_vs_summaries_vectors}.
Comparing the \taxtype{} probabilities in reviews and summaries, it is clear (and unsurprising) that summaries are considerably more efficient in communicating information. 
In particular, summaries predominantly describe the product, provide reasoning for opinions and offer many tips.
These findings are consistent with \citet{brazinskas2021amasum}, who describe the \emph{verdict} as emphasizing the most important points about the product, and the \emph{pros} and \emph{cons} as providing details about different aspects of the product.

Reviewers seem to add a non-negligible amount of personal information to the parts eventually excluded from summaries, possibly to make their reviews more relatable, for example, \say{We had used this book when my children were young and were identifying flowers.}
Interestingly, according to our analysis of helpfulness (Figure~\ref{fig_classification_vectors_hs}), readers seem to find this characteristic to be less helpful.
In general, we find (\autoref{tab_summary_helpfulness_correlation}) a much higher correlation between the vector of summaries and that of helpful sentences, than between the vector of summaries and that of unhelpful sentences.
In other words, good summaries tend to contain mainly helpful sentences. 
On the other hand, a helpful \textit{review} has a lower correlation with summaries, meaning that it still contains some potential verbosity, which is natural given the less stringent length constraints.

\begin{figure}
  \centering
  \includegraphics[width=0.75\columnwidth]{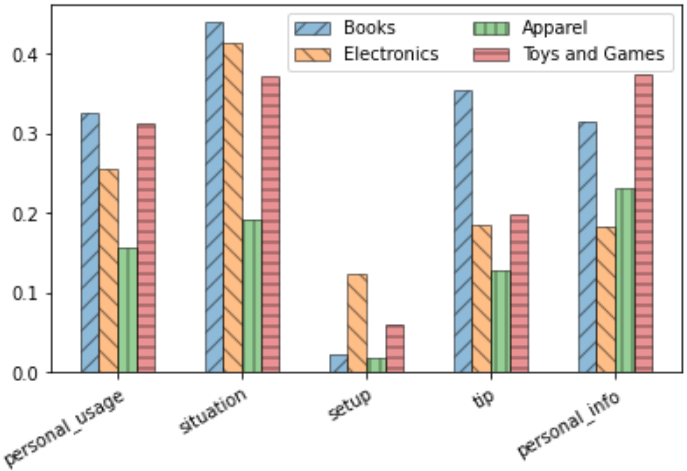}
  \caption{Aggregated review-level probabilities of select \taxtypes{}, grouped by product categories.}
  \label{fig_review_vectors_per_category}
\end{figure}

\subsection{Product Categories}
We conduct an examination of the reviews at the product \textit{category} level to learn what kind of information reviewers tend to share for different kinds of products.
\autoref{fig_review_vectors_per_category} reveals, for example, that people share much more \textit{personal information} when reviewing toys and games. 
We find that reviewers purchase these mostly for others, especially children, and therefore provide an explanation of who they bought the product for and their experience with it, e.g.,~\say{Hopefully my 3 year old doesn't toss all of the pieces everywhere.} 
A review is also, expectedly, more likely to discuss \textit{setup} of an electronics device than for other kinds of products. 
A book review is more likely to describe a \textit{situation} or provide a \textit{tip},
two \taxtypes{} we find often co-occurring within sentences suggesting usage for the book, e.g.,
\say{This book should probably be combined with other study methods to help your child perform to the best of their ability.}
We suggest that this behaviour may be specific to certain subcategories of books, such as self-help and other nonfiction. 
Overall, analyzing variability of \taxtypes{} across product categories allows data-driven discovery of insights about products and customers.

\subsection{Rhetorical Structure}
\label{sec_analysis_rhetorical}

Rhetorical Structure Theory studies the organization of a body of text and the relationship between its parts \citep{mann1988rst}, facilitating the assessment of a text's coherence. 
In our setting, we can examine how \taxtypes{} affect each other as a review or summary progresses. 
To demonstrate this, we compute the average \taxtype{} probabilities at each sentence position of reviews with 6 sentences, and summaries with 7 sentences (explained in Appendix~\ref{sec_appendix_experiments_analysis}), and plot them on a graph showing the progression.

\begin{figure}
  \centering
  \includegraphics[width=0.85\linewidth]{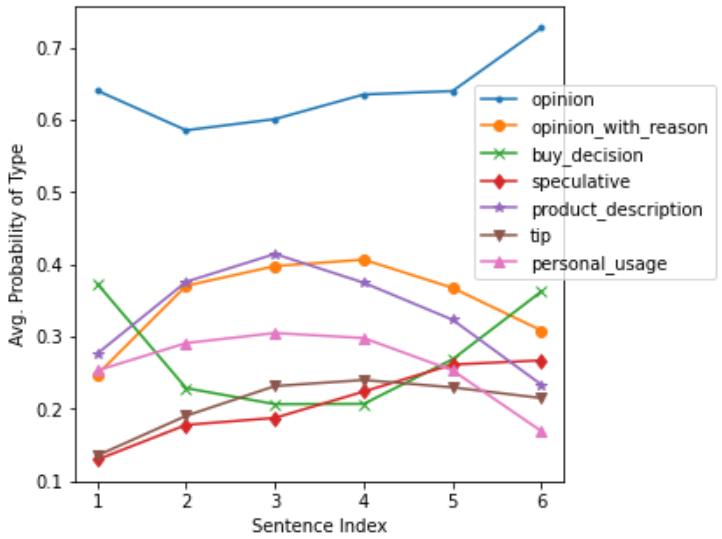}
  \caption{Expected probabilities of \taxtypes{} over the progression of a 6-sentence review in AmaSum.}
  \label{fig_review_structure}
\end{figure}

\begin{figure}
  \centering
  \includegraphics[width=0.85\linewidth]{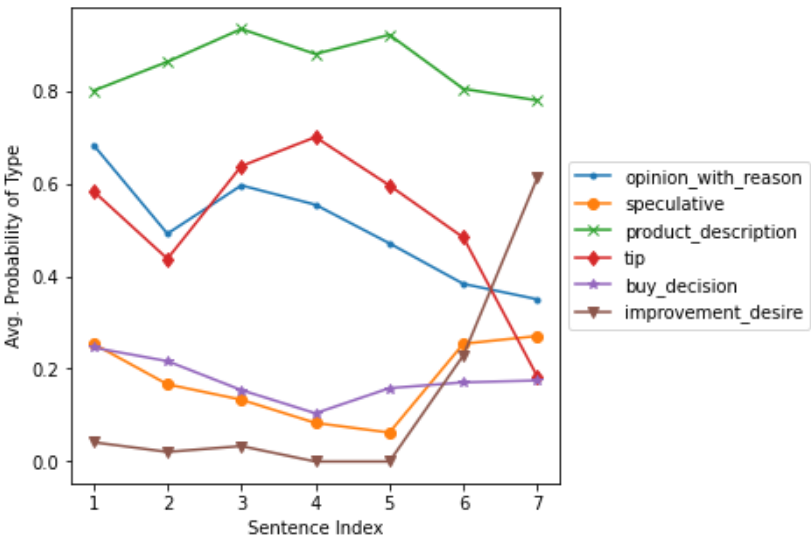}
  \caption{Expected probabilities of \taxtypes{} over the progression of a 7-sentence summary in AmaSum.}
  \label{fig_summary_structure}
\end{figure}

As can be seen in \autoref{fig_review_structure}, reviews tend to open and close with a \textit{buy decision} statement. 
Customers use such statements to introduce the situation, e.g.,~\say{Purchased this SSD in order to put Windows 7 and some games on it to decrease load times.}, or conclude with an emphasizing reiteration of satisfaction (or lack thereof), e.g.,~\say{I only regret not getting it sooner!}
Conversely, reasoned opinions and detailed information tend to occur in the middle of the review. 
We note that the likelihood of \textit{speculative} statements increases as the review progresses. 
Speculation should be delivered only after it is justified, e.g.,~\say{That being said, they're probably still ok for temporary storage.}

AmaSum summaries are formed by concatenating a \textit{verdict} (avg. 1.42 sentences in our data), a \textit{pros} section (avg. 4.04 sentences) and a \textit{cons} section (avg. 1.54 sentences). 
Due to this segmented configuration, the rhetorical structure is only partly relevant since global coherence is not an objective of the summary. 
\autoref{fig_summary_structure} demonstrates the behavior of the \emph{verdict} at the first 1--2 sentences, the next $\sim$4 sentences representing the structure of the \emph{pros}, and the last 1--2 corresponding to the \emph{cons} section. 
The \emph{verdict} embodies features that stand out most in the product with its overall sentiment, e.g., \say{Single camera setup that's reasonably priced because it doesn't require a separate base station.}
Indeed, the \textit{product description} and \textit{opinion with reason} \taxtypes{} are particularly likely. 
The \emph{pros} section generally describes advantages of the product ordered by importance, and accordingly we see reasoning and tip content peaking and then decreasing gradually along these sentences.
The \emph{cons} section provides increased  \textit{speculation} to suggest what the disadvantage might cause, e.g., \say{The 16-hour battery life can become frustrating for some}, as well as vastly increased incidence of \textit{improvement desire}, such as \say{Customers commented that these aren't absorbent enough to handle big messes.}

Future work may further examine review and summary structure through the lens of our typology to improve automatic review summarization and to control content type while maintaining coherence.



\section{Conclusion}
\label{sec_conclusion}
Product reviews contain a wealth of information that has been mined only shallowly over the years. 
We introduce a broad typology of information types that we show is quite useful for multiple popular downstream tasks on review data.
Thanks to an effective LLM-based predictor, our large-scale experiments and analyses can assist future endeavours to maximally benefit from the many available product reviews on the web. 
The revealed \taxtypes{} can also be used as-is for filtering information according to particular needs of customers or vendors, such as only showing descriptions of the product, or only suggesting improvements.

We advocate pursuing similar broad-ranged research on other domains, elaborating on our initial review-focused typology, enabling informed approaches for even more downstream applications.


\section*{Limitations}
The typology we suggest is partly inspired by previous related studies, and a result of a systematic examination of many review sentences which facilitated the accumulation of the typology types. There are likely other possible types and combinations of types that would establish an alternate typology. However, our research here mainly focuses on the use of a broad typology for revealing insights in product reviews and for assisting downstream applications.

In addition, our experiments and analyses rely on a predictor that does not label sentences at full accuracy. While our evaluation of the predictor gives us good reason to place confidence in its outputs, the conclusions we draw are certainly based on some noisy results. We therefore reported on findings from the analyses that are more distinctly evident from the data. Finally, when comparing or classifying between two subsets of texts (e.g., helpful vs. unhelpful, or reviews vs. summaries), the noise is analogously enforced on both subsets, making them generally comparable.

Lastly, our topology can inspire typologies in other review domains (such as hotels or dining) or other domains entirely. While some types in our typology are likely relevant and can be reused for other domains, developing this further requires a more thorough investigation in future work.



\bibliography{bibliography}

\begin{thebibliography}{46}
\expandafter\ifx\csname natexlab\endcsname\relax\def\natexlab#1{#1}\fi

\bibitem[{Adak et~al.(2022)Adak, Pradhan, and Shukla}]{Adak2022sentimentExplainability}
Anirban Adak, Biswajeet Pradhan, and Nagesh Shukla. 2022.
\newblock \href {https://doi.org/10.3390/foods11101500} {{Sentiment Analysis of Customer Reviews of Food Delivery Services Using Deep Learning and Explainable Artificial Intelligence: Systematic Review}}.
\newblock \emph{Foods}, 11(10).

\bibitem[{AI21labs(2023)}]{ai212023jurassic}
AI21labs. 2023.
\newblock \href {https://www.ai21.com/blog/introducing-j2} {{Announcing Jurassic-2 and Task-Specific APIs}}.
\newblock AI21 Labs Announcements.
\newblock Accessed: 2023-06-01.

\bibitem[{AI21Studio(2023)}]{ai212023pythonsdk}
AI21Studio. 2023.
\newblock \href {https://docs.ai21.com/reference/python-sdk} {{Python SDK}}.
\newblock AI21 Documentation.
\newblock Accessed: 2023-06-01.

\bibitem[{Ali and Hassan(2022)}]{ali2022framingSurvey}
Mohammad Ali and Naeemul Hassan. 2022.
\newblock \href {https://aclanthology.org/2022.emnlp-main.633} {{A Survey of Computational Framing Analysis Approaches}}.
\newblock In \emph{Proceedings of the 2022 Conference on Empirical Methods in Natural Language Processing}, pages 9335--9348, Abu Dhabi, United Arab Emirates. Association for Computational Linguistics.

\bibitem[{Bhattacharya et~al.(2020)Bhattacharya, Srinivas~N, Ramamurthy, Tharagam, and Narayana}]{Bhattacharya2020seller_delivery_classification}
Saikat Bhattacharya, Bharani Srinivas~N, Prakash Ramamurthy, Venkataramanan Tharagam, and D~Narayana. 2020.
\newblock \href {https://medium.com/@blogsupport/classification-of-product-reviews-in-e-commerce-website-and-proposing-a-balanced-rating-system-a1ebf6dd192c} {{Classification of product reviews in e-commerce website and proposing a balanced rating system}}.
\newblock medium.com.
\newblock Accessed: 2023-06-01.

\bibitem[{Bird(2006)}]{bird2006nltk}
Steven Bird. 2006.
\newblock \href {https://doi.org/10.3115/1225403.1225421} {{NLTK: The Natural Language Toolkit}}.
\newblock In \emph{Proceedings of the {COLING}/{ACL} 2006 Interactive Presentation Sessions}, pages 69--72, Sydney, Australia. Association for Computational Linguistics.

\bibitem[{Bra{\v{z}}inskas et~al.(2021)Bra{\v{z}}inskas, Lapata, and Titov}]{brazinskas2021amasum}
Arthur Bra{\v{z}}inskas, Mirella Lapata, and Ivan Titov. 2021.
\newblock \href {https://doi.org/10.18653/v1/2021.emnlp-main.743} {{Learning Opinion Summarizers by Selecting Informative Reviews}}.
\newblock In \emph{Proceedings of the 2021 Conference on Empirical Methods in Natural Language Processing}, pages 9424--9442, Online and Punta Cana, Dominican Republic. Association for Computational Linguistics.

\bibitem[{Chen et~al.(2017)Chen, Xu, He, and Wang}]{chen2017sentAnaWithTypeClassification}
Tao Chen, Ruifeng Xu, Yulan He, and Xuan Wang. 2017.
\newblock \href {https://doi.org/https://doi.org/10.1016/j.eswa.2016.10.065} {{Improving Sentiment Analysis via Sentence Type Classification using BiLSTM-CRF and CNN}}.
\newblock \emph{Expert Systems with Applications}, 72:221--230.

\bibitem[{Chen et~al.(2022)Chen, Verdi~do Amarante, Donaldson, Jo, and Park}]{chen2022argumentMiningForHelpfulness}
Zaiqian Chen, Daniel Verdi~do Amarante, Jenna Donaldson, Yohan Jo, and Joonsuk Park. 2022.
\newblock \href {https://aclanthology.org/2022.emnlp-main.609} {{Argument Mining for Review Helpfulness Prediction}}.
\newblock In \emph{Proceedings of the 2022 Conference on Empirical Methods in Natural Language Processing}, pages 8914--8922, Abu Dhabi, United Arab Emirates. Association for Computational Linguistics.

\bibitem[{Chung et~al.(2022)Chung, Hou, Longpre, Zoph, Tay, Fedus, Li, Wang, Dehghani, Brahma, Webson, Gu, Dai, Suzgun, Chen, Chowdhery, Castro-Ros, Pellat, Robinson, Valter, Narang, Mishra, Yu, Zhao, Huang, Dai, Yu, Petrov, Chi, Dean, Devlin, Roberts, Zhou, Le, and Wei}]{chung2022flan}
Hyung~Won Chung, Le~Hou, Shayne Longpre, Barret Zoph, Yi~Tay, William Fedus, Yunxuan Li, Xuezhi Wang, Mostafa Dehghani, Siddhartha Brahma, Albert Webson, Shixiang~Shane Gu, Zhuyun Dai, Mirac Suzgun, Xinyun Chen, Aakanksha Chowdhery, Alex Castro-Ros, Marie Pellat, Kevin Robinson, Dasha Valter, Sharan Narang, Gaurav Mishra, Adams Yu, Vincent Zhao, Yanping Huang, Andrew Dai, Hongkun Yu, Slav Petrov, Ed~H. Chi, Jeff Dean, Jacob Devlin, Adam Roberts, Denny Zhou, Quoc~V. Le, and Jason Wei. 2022.
\newblock \href {http://arxiv.org/abs/2210.11416} {{Scaling Instruction-Finetuned Language Models}}.

\bibitem[{Cohan et~al.(2019)Cohan, Beltagy, King, Dalvi, and Weld}]{cohan2019articleClassification}
Arman Cohan, Iz~Beltagy, Daniel King, Bhavana Dalvi, and Dan Weld. 2019.
\newblock \href {https://doi.org/10.18653/v1/D19-1383} {{Pretrained Language Models for Sequential Sentence Classification}}.
\newblock In \emph{Proceedings of the 2019 Conference on Empirical Methods in Natural Language Processing and the 9th International Joint Conference on Natural Language Processing (EMNLP-IJCNLP)}, pages 3693--3699, Hong Kong, China. Association for Computational Linguistics.

\bibitem[{Cortes and Vapnik(1995)}]{Cortes1995svm}
Corinna Cortes and Vladimir Vapnik. 1995.
\newblock \href {https://doi.org/10.1007/BF00994018} {{Support-Vector Networks}}.
\newblock \emph{Machine Learning}, 20(3):273–297.

\bibitem[{Dernoncourt and Lee(2017)}]{dernoncourt2017pubmed}
Franck Dernoncourt and Ji~Young Lee. 2017.
\newblock \href {https://aclanthology.org/I17-2052} {{PubMed 200k RCT: a Dataset for Sequential Sentence Classification in Medical Abstracts}}.
\newblock In \emph{Proceedings of the Eighth International Joint Conference on Natural Language Processing (Volume 2: Short Papers)}, pages 308--313, Taipei, Taiwan. Asian Federation of Natural Language Processing.

\bibitem[{Desmet and Fokkinga(2020)}]{Desmet2020psychoTypology}
Pieter Desmet and Steven Fokkinga. 2020.
\newblock \href {https://doi.org/10.3390/mti4030038} {{Beyond Maslow’s Pyramid: Introducing a Typology of Thirteen Fundamental Needs for Human-Centered Design}}.
\newblock \emph{Multimodal Technologies and Interaction}, 4(3).

\bibitem[{Djuric et~al.(2015)Djuric, Zhou, Morris, Grbovic, Radosavljevic, and Bhamidipati}]{Djuric2015hatespeech}
Nemanja Djuric, Jing Zhou, Robin Morris, Mihajlo Grbovic, Vladan Radosavljevic, and Narayan Bhamidipati. 2015.
\newblock \href {https://doi.org/10.1145/2740908.2742760} {{Hate Speech Detection with Comment Embeddings}}.
\newblock In \emph{Proceedings of the 24th International Conference on World Wide Web}, WWW '15 Companion, page 29–30, New York, NY, USA. Association for Computing Machinery.

\bibitem[{Eckle-Kohler et~al.(2015)Eckle-Kohler, Kluge, and Gurevych}]{ecklekohler2015arguments}
Judith Eckle-Kohler, Roland Kluge, and Iryna Gurevych. 2015.
\newblock \href {https://doi.org/10.18653/v1/D15-1267} {{On the Role of Discourse Markers for Discriminating Claims and Premises in Argumentative Discourse}}.
\newblock In \emph{Proceedings of the 2015 Conference on Empirical Methods in Natural Language Processing}, pages 2236--2242, Lisbon, Portugal. Association for Computational Linguistics.

\bibitem[{Efron(1992)}]{efron1992bootstrap}
Bradley Efron. 1992.
\newblock \href {https://doi.org/10.1007/978-1-4612-4380-9_41} {{Bootstrap Methods: Another Look at the Jackknife}}.
\newblock In Samuel Kotz and Norman~L. Johnson, editors, \emph{Breakthroughs in Statistics: Methodology and Distribution}, pages 569--593. Springer New York, New York, NY.

\bibitem[{Farber et~al.(2022)Farber, Carmel, Kuchy, and Mejer}]{farber2022tips}
Miriam Farber, David Carmel, Lital Kuchy, and Avihai Mejer. 2022.
\newblock \href {https://doi.org/10.1145/3477495.3531805} {{Analyzing the Support Level for Tips Extracted from Product Reviews}}.
\newblock In \emph{Proceedings of the 45th International ACM SIGIR Conference on Research and Development in Information Retrieval}, SIGIR '22, page 2059–2064, New York, NY, USA. Association for Computing Machinery.

\bibitem[{Feine et~al.(2019)Feine, Gnewuch, Morana, and Maedche}]{Feine2019conversationTaxonomy}
Jasper Feine, Ulrich Gnewuch, Stefan Morana, and Alexander Maedche. 2019.
\newblock \href {https://doi.org/https://doi.org/10.1016/j.ijhcs.2019.07.009} {{A Taxonomy of Social Cues for Conversational Agents}}.
\newblock \emph{International Journal of Human-Computer Studies}, 132:138--161.

\bibitem[{Gamzu et~al.(2021)Gamzu, Gonen, Kutiel, Levy, and Agichtein}]{gamzu2021helpfulsentences}
Iftah Gamzu, Hila Gonen, Gilad Kutiel, Ran Levy, and Eugene Agichtein. 2021.
\newblock \href {https://doi.org/10.18653/v1/2021.naacl-main.55} {{Identifying Helpful Sentences in Product Reviews}}.
\newblock In \emph{Proceedings of the 2021 Conference of the North American Chapter of the Association for Computational Linguistics: Human Language Technologies}, pages 678--691, Online. Association for Computational Linguistics.

\bibitem[{He and McAuley(2016)}]{he2016reviewsDS}
Ruining He and Julian McAuley. 2016.
\newblock \href {https://doi.org/10.1145/2872427.2883037} {{Ups and Downs: Modeling the Visual Evolution of Fashion Trends with One-Class Collaborative Filtering}}.
\newblock In \emph{Proceedings of the 25th International Conference on World Wide Web}, WWW '16, page 507–517, Republic and Canton of Geneva, CHE. International World Wide Web Conferences Steering Committee.

\bibitem[{Hirsch et~al.(2021)Hirsch, Novgorodov, Guy, and Nus}]{hirsch2021producttips}
Sharon Hirsch, Slava Novgorodov, Ido Guy, and Alexander Nus. 2021.
\newblock \href {https://doi.org/10.1145/3437963.3441755} {{Generating Tips from Product Reviews}}.
\newblock In \emph{Proceedings of the 14th ACM International Conference on Web Search and Data Mining}, WSDM '21, page 310–318, New York, NY, USA. Association for Computing Machinery.

\bibitem[{Joshi et~al.(2017)Joshi, Bhattacharyya, and Carman}]{aditya2017sarcasm}
Aditya Joshi, Pushpak Bhattacharyya, and Mark~J. Carman. 2017.
\newblock \href {https://doi.org/10.1145/3124420} {{Automatic Sarcasm Detection: A Survey}}.
\newblock \emph{ACM Comput. Surv.}, 50(5).

\bibitem[{Levy(1979)}]{levy1979CommunicativeGoals}
David~M. Levy. 1979.
\newblock \href {https://doi.org/https://doi.org/10.1163/9789004368897_009} {\emph{{Communicative Goals and Strategies: Between Discourse and Syntax}}}, Syntax and Semantics, pages 183 -- 210. Brill, Leiden, The Netherlands.

\bibitem[{Liu et~al.(2017)Liu, Gao, Lv, Li, Geng, Li, and Wang}]{liu2017argsForHelpfullness}
Haijing Liu, Yang Gao, Pin Lv, Mengxue Li, Shiqiang Geng, Minglan Li, and Hao Wang. 2017.
\newblock \href {https://doi.org/10.18653/v1/D17-1142} {{Using Argument-based Features to Predict and Analyse Review Helpfulness}}.
\newblock In \emph{Proceedings of the 2017 Conference on Empirical Methods in Natural Language Processing}, pages 1358--1363, Copenhagen, Denmark. Association for Computational Linguistics.

\bibitem[{Liu et~al.(2023)Liu, Yuan, Fu, Jiang, Hayashi, and Neubig}]{liu2023prompt}
Pengfei Liu, Weizhe Yuan, Jinlan Fu, Zhengbao Jiang, Hiroaki Hayashi, and Graham Neubig. 2023.
\newblock \href {https://doi.org/10.1145/3560815} {{Pre-Train, Prompt, and Predict: A Systematic Survey of Prompting Methods in Natural Language Processing}}.
\newblock \emph{ACM Comput. Surv.}, 55(9).

\bibitem[{Louis and Nenkova(2014)}]{louis2014summaryContent}
Annie Louis and Ani Nenkova. 2014.
\newblock \href {https://doi.org/10.3115/v1/E14-1067} {{Verbose, Laconic or Just Right: A Simple Computational Model of Content Appropriateness under Length Constraints}}.
\newblock In \emph{Proceedings of the 14th Conference of the {E}uropean Chapter of the Association for Computational Linguistics}, pages 636--644, Gothenburg, Sweden. Association for Computational Linguistics.

\bibitem[{Mann and Thompson(1988)}]{mann1988rst}
William~C. Mann and Sandra~A. Thompson. 1988.
\newblock \href {https://doi.org/doi:10.1515/text.1.1988.8.3.243} {{Rhetorical Structure Theory: Toward a functional theory of text organization}}.
\newblock \emph{Text - Interdisciplinary Journal for the Study of Discourse}, 8(3):243--281.

\bibitem[{Naderi and Hirst(2017)}]{naderi2017newsFrames}
Nona Naderi and Graeme Hirst. 2017.
\newblock \href {https://doi.org/10.26615/978-954-452-049-6_070} {{Classifying Frames at the Sentence Level in News Articles}}.
\newblock In \emph{Proceedings of the International Conference Recent Advances in Natural Language Processing, {RANLP} 2017}, pages 536--542, Varna, Bulgaria. INCOMA Ltd.

\bibitem[{Nickerson et~al.(2013)Nickerson, Varshney, and Muntermann}]{Nickerson2013taxonomydevelop}
Robert~C Nickerson, Upkar Varshney, and Jan Muntermann. 2013.
\newblock \href {https://doi.org/10.1057/ejis.2012.26} {{A method for taxonomy development and its application in information systems}}.
\newblock \emph{European Journal of Information Systems}, 22(3):336--359.

\bibitem[{Ocampo~Diaz and Ng(2018)}]{ocampodiaz2018helpfulnessFeatures}
Gerardo Ocampo~Diaz and Vincent Ng. 2018.
\newblock \href {https://doi.org/10.18653/v1/P18-1065} {{Modeling and Prediction of Online Product Review Helpfulness: A Survey}}.
\newblock In \emph{Proceedings of the 56th Annual Meeting of the Association for Computational Linguistics (Volume 1: Long Papers)}, pages 698--708, Melbourne, Australia. Association for Computational Linguistics.

\bibitem[{Passon et~al.(2018)Passon, Lippi, Serra, and Tasso}]{passon2018helpfulness}
Marco Passon, Marco Lippi, Giuseppe Serra, and Carlo Tasso. 2018.
\newblock \href {https://doi.org/10.18653/v1/W18-5205} {{Predicting the Usefulness of Amazon Reviews Using Off-The-Shelf Argumentation Mining}}.
\newblock In \emph{Proceedings of the 5th Workshop on Argument Mining}, pages 35--39, Brussels, Belgium. Association for Computational Linguistics.

\bibitem[{Rohrdantz(2016)}]{Rohrdantz2016review_insights}
Christian Rohrdantz. 2016.
\newblock \href {https://www.linkedin.com/pulse/taxonomy-consumer-insights-from-customer-reviews-christian-rohrdantz/} {{Taxonomy of Consumer Insights from Customer Reviews}}.
\newblock linkedin.com.
\newblock Accessed: 2023-06-01.

\bibitem[{Serrano-Guerrero et~al.(2015)Serrano-Guerrero, Olivas, Romero, and Herrera-Viedma}]{serranoguerrero2015sentAna}
Jesus Serrano-Guerrero, Jose~A. Olivas, Francisco~P. Romero, and Enrique Herrera-Viedma. 2015.
\newblock \href {https://doi.org/https://doi.org/10.1016/j.ins.2015.03.040} {{Sentiment analysis: A review and comparative analysis of web services}}.
\newblock \emph{Information Sciences}, 311:18--38.

\bibitem[{Shivaprasad and Shetty(2017)}]{Shivaprasad2017SentAnaRev}
T.~K. Shivaprasad and Jyothi Shetty. 2017.
\newblock \href {https://doi.org/10.1109/ICICCT.2017.7975207} {{Sentiment Analysis of Product Reviews: A Review}}.
\newblock In \emph{2017 International Conference on Inventive Communication and Computational Technologies (ICICCT)}, pages 298--301.

\bibitem[{Tay et~al.(2023)Tay, Dehghani, Tran, Garcia, Wei, Wang, Chung, Shakeri, Bahri, Schuster, Zheng, Zhou, Houlsby, and Metzler}]{tay2023ul2}
Yi~Tay, Mostafa Dehghani, Vinh~Q. Tran, Xavier Garcia, Jason Wei, Xuezhi Wang, Hyung~Won Chung, Siamak Shakeri, Dara Bahri, Tal Schuster, Huaixiu~Steven Zheng, Denny Zhou, Neil Houlsby, and Donald Metzler. 2023.
\newblock \href {http://arxiv.org/abs/2205.05131} {{UL2: Unifying Language Learning Paradigms}}.

\bibitem[{Tesfagergish et~al.(2022)Tesfagergish, Kapočiūtė-Dzikienė, and Damaševičius}]{Tesfagergish2022zeroshot_classification}
Senait~Gebremichael Tesfagergish, Jurgita Kapočiūtė-Dzikienė, and Robertas Damaševičius. 2022.
\newblock \href {https://doi.org/10.3390/app12178662} {{Zero-Shot Emotion Detection for Semi-Supervised Sentiment Analysis Using Sentence Transformers and Ensemble Learning}}.
\newblock \emph{Applied Sciences}, 12(17).

\bibitem[{Tubishat et~al.(2018)Tubishat, Idris, and Abushariah}]{Tubishat2018aspExtr}
Mohammad Tubishat, Norisma Idris, and Mohammad~A.M. Abushariah. 2018.
\newblock \href {https://doi.org/https://doi.org/10.1016/j.ipm.2018.03.008} {{Implicit aspect extraction in sentiment analysis: Review, taxonomy, oppportunities, and open challenges}}.
\newblock \emph{Information Processing \& Management}, 54(4):545--563.

\bibitem[{Vinodhini and Chandrasekaran(2012)}]{vinodhini2012opinionmining}
G~Vinodhini and RM~Chandrasekaran. 2012.
\newblock \href {https://www.researchgate.net/profile/Vinodhini-G-2/publication/265163299_Sentiment_Analysis_and_Opinion_Mining_A_Survey/links/54018f330cf2bba34c1af133/Sentiment-Analysis-and-Opinion-Mining-A-Survey.pdf} {{Sentiment Analysis and Opinion Mining: A Survey}}.
\newblock \emph{International Journal of Advanced Research in Computer Science and Software Engineering}, 2(6).

\bibitem[{Wei et~al.(2022{\natexlab{a}})Wei, Bosma, Zhao, Guu, Yu, Lester, Du, Dai, and Le}]{wei2022flan}
Jason Wei, Maarten Bosma, Vincent Zhao, Kelvin Guu, Adams~Wei Yu, Brian Lester, Nan Du, Andrew~M. Dai, and Quoc~V Le. 2022{\natexlab{a}}.
\newblock \href {https://openreview.net/forum?id=gEZrGCozdqR} {{Finetuned Language Models are Zero-Shot Learners}}.
\newblock In \emph{International Conference on Learning Representations}.

\bibitem[{Wei et~al.(2022{\natexlab{b}})Wei, Wang, Schuurmans, Bosma, brian ichter, Xia, Chi, Le, and Zhou}]{wei2022chainofthought}
Jason Wei, Xuezhi Wang, Dale Schuurmans, Maarten Bosma, brian ichter, Fei Xia, Ed~H. Chi, Quoc~V Le, and Denny Zhou. 2022{\natexlab{b}}.
\newblock \href {https://openreview.net/forum?id=_VjQlMeSB_J} {{Chain of Thought Prompting Elicits Reasoning in Large Language Models}}.
\newblock In \emph{Advances in Neural Information Processing Systems}.

\bibitem[{Wolf et~al.(2020)Wolf, Debut, Sanh, Chaumond, Delangue, Moi, Cistac, Rault, Louf, Funtowicz, Davison, Shleifer, von Platen, Ma, Jernite, Plu, Xu, Le~Scao, Gugger, Drame, Lhoest, and Rush}]{wolf2020huggingface}
Thomas Wolf, Lysandre Debut, Victor Sanh, Julien Chaumond, Clement Delangue, Anthony Moi, Pierric Cistac, Tim Rault, Remi Louf, Morgan Funtowicz, Joe Davison, Sam Shleifer, Patrick von Platen, Clara Ma, Yacine Jernite, Julien Plu, Canwen Xu, Teven Le~Scao, Sylvain Gugger, Mariama Drame, Quentin Lhoest, and Alexander Rush. 2020.
\newblock \href {https://doi.org/10.18653/v1/2020.emnlp-demos.6} {{Transformers: State-of-the-Art Natural Language Processing}}.
\newblock In \emph{Proceedings of the 2020 Conference on Empirical Methods in Natural Language Processing: System Demonstrations}, pages 38--45, Online. Association for Computational Linguistics.

\bibitem[{Xu et~al.(2019)Xu, Liu, Shu, and Yu}]{xu2019bertSentAna}
Hu~Xu, Bing Liu, Lei Shu, and Philip Yu. 2019.
\newblock \href {https://doi.org/10.18653/v1/N19-1242} {{BERT Post-Training for Review Reading Comprehension and Aspect-based Sentiment Analysis}}.
\newblock In \emph{Proceedings of the 2019 Conference of the North {A}merican Chapter of the Association for Computational Linguistics: Human Language Technologies, Volume 1 (Long and Short Papers)}, pages 2324--2335, Minneapolis, Minnesota. Association for Computational Linguistics.

\bibitem[{Xu and Liang(2001)}]{xu2001crossVal}
Qing-Song Xu and Yi-Zeng Liang. 2001.
\newblock \href {https://doi.org/https://doi.org/10.1016/S0169-7439(00)00122-2} {{Monte Carlo Cross Validation}}.
\newblock \emph{Chemometrics and Intelligent Laboratory Systems}, 56(1):1--11.

\bibitem[{Yadav and Vishwakarma(2020)}]{Yadav2020sentAna}
Ashima Yadav and Dinesh~Kumar Vishwakarma. 2020.
\newblock \href {https://doi.org/10.1007/s10462-019-09794-5} {{Sentiment analysis using deep learning architectures: a review}}.
\newblock \emph{Artificial Intelligence Review}, 53(6):4335--4385.

\bibitem[{Youden(1950)}]{youden1950youdenj}
W.~J. Youden. 1950.
\newblock \href {https://doi.org/https://doi.org/10.1002/1097-0142(1950)3:1<32::AID-CNCR2820030106>3.0.CO;2-3} {{Index for Rating Diagnostic Tests}}.
\newblock \emph{Cancer}, 3(1):32--35.

\end{thebibliography}

\newpage

\appendix

\section{Typology Details}
\label{sec_appendix_taxonomy}

The 8 categories from which the sentences were taken initially for deriving \taxtypes{} are ``Fashion'', ``Automotive'', ``Books'', ``CDs and Vinyls'', ``Digital Music'', ``Electronics'', ``Movies and TV'', ``Toys and Games''. These were chosen due to the expected differences in aspects associated to the categories, to allow variability and ensure an exhaustive view at types found within reviews.

The \textit{inappropriate} \taxtype{} was not found in the initial data we used, likely because e-commerce websites make an effort to remove such reviews. However this \taxtype{} is a know characteristic in reviews data, and sometimes seeps in anyways.

We used NLTK \citep{bird2006nltk} sentence tokenization to split reviews to sentences, and applied simple normalization techniques to clean the texts. This procedure was used also for the rest of the experiments and analyses in the paper.

\section{Predictor Model Details}
\label{sec_appendix_predictor}

\subsection{Development and Test Evaluation}
\label{sec_appendix_predictor_data_evaluation}


\paragraph{Development evaluation.}
All our initial assessments on the models were done on the development set. We tried several temperatures and engineered some prompts for the models. We only evaluated models that all its responses started with a \say{yes} or \say{no}. 

We extracted \textbf{optimal thresholds} for all \taxtypes{} as follows: For thresholds $[0.1, 1.0]$ incrementing by $0.1$, and for each \taxtype{}, we computed the $F_1$ score against the annotations. The threshold with the maximum $F_1$ score was chosen as the optimal threshold for the corresponding \taxtype.

The macro-$F_1$ score for a model is then the average of $F_1$ scores over all \taxtypes{} with their optimal thresholds. The main results on the development set are presented in \autoref{tab_dev_results}.

\begin{table}[b]
    \centering
    \resizebox{\columnwidth}{!}{
    \begin{tabular}{lcccc}
        \toprule
        \multirow{2}{*}{\makecell{Model}} &
        \multirow{2}{*}{\makecell{Temperature}} &
        \multirow{2}{*}{\makecell{\# Repetitions\\Per Prompt}} &
        \multirow{2}{*}{\makecell{Macro \\ $F_1$}} &
        \multirow{2}{*}{\makecell{Run Time\\Per Sentence}} \\
        & & & & \\
        \midrule
        \texttt{flan-t5-xxl} & $0.7$ & 10 & $41.7$ & $\sim 10.8$ sec. \\
        \texttt{flan-t5-xxl} & $0.3$ & 10 & $56.9$ & $\sim 10.8$ sec. \\
        \texttt{flan-t5-xxl} & $0.3$ & 30 & $57.2$ & $\sim 10.8$ sec. \\
        \texttt{flan-ul2}    & $0.3$ & 10 & $51.5$ & $\sim 21.0$ sec. \\
        \texttt{flan-ul2}    & $0.3$ & 30 & $52.2$ & $\sim 21.0$ sec. \\
        \bottomrule
    \end{tabular}}
    \caption{The results on the \textit{development} set of the main models in the initial assessment. The Macro $F_1$ score is computed based on optimal thresholds for each of the \taxtypes{}.}
    \label{tab_dev_results}
\end{table}

Using 30 repetitions per prompt only slightly improves results, which is not worth the increased compute resources.

A supplementary manual inspection of the predictions gave reason to rely confidently enough on the \texttt{flan-t5-xxl} model. In fact, we noticed that many of the \taxtypes{} predicted by the models, that were not labeled in the development set, were indeed valid in certain contexts. In any case, our aim is not to produce the best model, but to rely on one enough for our analyses.

We note that all other models we tried gave substantially worse results, or irrelevant responses altogether. On the other hand, the models presented in \autoref{tab_dev_results} were very consistent both in output format and final results.


\paragraph{Test evaluation.}
We set the thresholds for all \taxtypes{} with those computed on the development set. The $F_1$ score was then computed for each \taxtype{} on the test set, and the average of those produces the final macro-$F_1$ score, as presented in \autoref{tab_test_scores_per_type}. The recall and precision indicate a finer-grained view of the capabilities of the predictor. On manual inspection, many of the incorrectly labeled sentences could be correct in certain contexts. I.e., when the predictor labels incorrectly, it is not nonsensically wrong.

\subsection{Model}
\label{sec_appendix_predictor_model}

\paragraph{Model implementations.}
All FLAN models were run using the Huggingface transformers library \citep{wolf2020huggingface}. The Jurassic model was run using the AI21 Python SDK \citep{ai212023pythonsdk} with a very small monetary budget. In addition to \texttt{flan-t5-\{xl, xxl\}}, \texttt{flan-ul2} and \texttt{j2-jumbo-instruct}, we also tested \texttt{flan-t5-\{base, large\}}, which produced considerably poorer outputs.

\paragraph{Temperature.}
On the models assessed, we mainly tested temperatures of $0.3$ and $0.7$. The FLAN models worked better with $0.3$ and the Jurassic model worked better with $0.7$ (as evaluated on the development set). However, as said, only the \texttt{flan-t5-xxl} and \texttt{flan-ul2} produced reasonable enough results in any case.

\paragraph{Hardware.}
We ran all FLAN predictions on an AWS \texttt{g4dn.12xlarge} EC2 server, which includes 4 NVIDIA T4 GPUs, 64 GB GPU memory and 192 GB RAM. The Jurassic model runs through an API.

\paragraph{Run time.}
The \texttt{flan-t5-xxl} model's run time was about $0.45$ seconds per prompt (including the 10 repetitions for computing probability). Hence, each sentence required an average of $10.8$ seconds for full typology prediction.

\paragraph{Prompt.}
The chosen prompt used in the \texttt{flan-t5-xxl} model is:

\begin{table}[h]
    \vspace{-2mm}
    \centering
    \resizebox{\columnwidth}{!}{
    \begin{tabular}{|l|}
        \hline
        \textsl{
        \begin{tabular}[c]{@{}l@{}}`Given that this sentence is from a product review about\\\texttt{\{product category subprompt\}}, \texttt{\{}\taxtype{} \texttt{question\}}?\\Answer yes or no. The sentence is: \say{\texttt{\{sentence\}}}'\end{tabular}} \\
        \hline
    \end{tabular}}
    \vspace{-4mm}
\end{table}

The \texttt{product category subprompt} is a string prepared for each category. For example the subprompt for the `Electronics' category would be ``\textit{an electronics product}'', and for `Toys and Games' it would be ``\textit{a toy or game}''.

The \texttt{\taxtype{} question} prompt substring is a question prepared for each \taxtype{}, as found in \autoref{tab_taxonomy_types}.

An example of a prompt for a sentence from the ``Books'' category for the \textit{product usage} \taxtype{} would be:

\begin{table}[h]
    \vspace{-2mm}
    \centering
    \resizebox{\columnwidth}{!}{
    \begin{tabular}{|l|}
        \hline
        \textsl{
        \begin{tabular}[c]{@{}l@{}}`Given that this sentence is from a product review about\\a book, does the sentence describe how the product can\\be used? Answer yes or no. The sentence is: ``I love this\\book and just  completed an incredible weekend workshop\\with the author.'''\end{tabular}} \\
        \hline
    \end{tabular}}
\end{table}

The \taxtype{} predictions on which we rely throughout this paper can be reproduced using these prompts with the \texttt{flan-t5-xxl} Huggingface model. Complementing code will be available as well.

\section{Downstream Task Experiment Details}
\label{sec_appendix_experiments}

This appendix describes the data, training, testing and evaluation details in all the experiments throughout the paper.

\subsection{Prediction of Review Helpfulness}
\label{sec_appendix_experiments_review_helpfulness}

\begin{table}[t]
    \centering
    \resizebox{\columnwidth}{!}{
    \begin{tabular}{lccc}
        \toprule
                 & \multirow{2}{*}{\makecell{Helpful\\Reviews}} & \multirow{2}{*}{\makecell{Helpful\\Sentences}} & \multirow{2}{*}{\makecell{Sentiment\\Analysis}} \\
        \taxtype{} Set & & & \\
        \midrule
        subjective      & 69.6 & 82.1 & 83.7 \\
        ~~~opinions      & 66.6 & 80.8 & 80.2 \\
        ~~~~~~(\textit{opinion} only)    & 48.7 & 72.7 & 80.2 \\
        ~~~~~~(\textit{op. w/rsn.} only) & 67.4 & 80.9 & 80.2 \\
        objective       & 67.3 & 83.3 & 80.2 \\
        ~~~description & 66.6 & 81.9 & 80.2 \\
        ~~~comparisons & 51.4 & 57.9 & 80.1 \\
        personal        & 60.3 & 64.0 & 80.1 \\
        non-product     & 56.8 & 70.4 & 80.1 \\
        stylistic       & 63.1 & 63.7 & 84.0 \\
        \midrule
        All             & 72.6 & 88.3 & 88.1 \\
        \bottomrule
    \end{tabular}}
    \caption{Accuracy scores (\%) on the classification tasks using an SVM with different subsets of \taxtypes{} from the taxonomy (\autoref{tab_taxonomy_groups}). An indented set is contained in the set hierarchically above it. This is a full version extending \autoref{tab_classification_results}.}
    \label{tab_classification_results_full}
\end{table}

\paragraph{Data used.}
\citet{gamzu2021helpfulsentences} released a dataset of helpful sentences, linked to their full review data. The data is based on 123 products from categories ``Toys'', ``Books'', ``Movies'', ``Music'', ``Camera'' and ``Electronics''. We targeted an approximately balanced amount of helpful to unhelpful reviews (regardless of the sentence-level helpfulness which \citeauthor{gamzu2021helpfulsentences} target). Each review has `helpful-count' and `nothelpful-count' fields. By setting helpful-count $>= 9$ and nothelpful-count $== 0$ for helpful reviews, and nothelpful-count $>= 3$ and helpful-count $== 0$ for unhelpful reviews, we produce 458 and 486 reviews respectively. The aim was to collect about 500 reviews for each class.

\paragraph{Training and testing.}
We use 50 iterations of cross validation, with a train set of 70\% and test if 30\%. The procedure is repeated for the different \taxtype{} sets separately.

\paragraph{Evaluation.}
As common in binary classification tasks, we report the accuracy measure \citep{Yadav2020sentAna}.

\paragraph{More results.}
\autoref{tab_classification_results_full} extends \autoref{tab_classification_results}, including all \taxtype{} sets defined in \autoref{tab_taxonomy_groups}.

\subsection{Prediction of Review Sentence Helpfulness}
\label{sec_appendix_experiments_sentence_helpfulness}

\begin{table}[t]
    \centering
    \resizebox{\columnwidth}{!}{
    \begin{tabular}{clccc}
        \toprule
        \multicolumn{2}{l}{Method} & MSE $
        \downarrow$ & PC $\uparrow$ & N@1 $\uparrow$ \\
        \midrule
        \multirow{11}{*}{\rotatebox[origin=c]{90}{{\small Taxonomy \taxtype{} Set w/ Log. Regr.}}}
        & subjective      & 0.096 & 0.66 & 0.86  \\
        & ~~~opinions        & 0.101 & 0.64 & 0.75  \\
        & ~~~~~~(\textit{opinion} only)    & 0.121 & 0.53 & 0.66  \\
        & ~~~~~~(\textit{op. w/rsn.} only) & 0.116 & 0.56 & 0.74  \\
        & objective       & 0.103 & 0.63 & 0.86  \\
        & ~~~description     & 0.110 & 0.59 & 0.86  \\
        & ~~~comparisons     & 0.165 & 0.08 & 0.69  \\
        & personal        & 0.152 & 0.31 & 0.62  \\
        & non-product     & 0.139 & 0.42 & 0.64  \\
        & stylistic      & 0.149 & 0.33 & 0.57  \\
        \cmidrule{2-5}
        & All             & 0.075 & 0.75 & 0.86  \\
        \midrule
        \multirow{4}{*}{\rotatebox[origin=c]{90}{{\small\citeauthor{gamzu2021helpfulsentences}}}} & Random          & 0.500 & 0.02 & 0.68 \\
        & Baseline 1 {\small(TF-IDF)}        & 0.090 & 0.63 & 0.91 \\
        & Baseline 2 {\small(ST-RIDGE)}        & 0.062 & 0.78 & 0.94 \\
        & Best {\small(BERT)}            & 0.053 & 0.84 & 0.95 \\
        \bottomrule
    \end{tabular}}
    \caption{Mean squared error, Pearson correlation and NDCG@1 scores on the helpful sentence scoring task using Logistic Regression with different subsets of \taxtypes{} from the taxonomy (\autoref{tab_taxonomy_groups}), and compared to results of \citet{gamzu2021helpfulsentences}. An indented set is contained in the set hierarchically above it. This is a full version extending \autoref{tab_helpful_sentences_results}.}
    \label{tab_helpful_sentences_results_full}
\end{table}

\paragraph{Data used.}
The full data of \citet{gamzu2021helpfulsentences} consists of 20000 sentences in the train set and 2000 in the test set. Each sentence has a continuous helpfulness score between 0 and 2. In addition, we find the helpfulness scores marking the borders of the top and bottom tertiles in the train set ($1.4$ and $1.0$), and mark the sentences as helpful or unhelpful respectively, or neutral for the mid-section. We use the same border scores to mark the sentences in the test set. Overall there are (7475, 7072, 5453) (unhelpful, helpful, neutral) sentences in the train, and (742, 562, 696) in the test. Notice that classes are not perfectly balanced in the train set since the scores on the borders (1.4 and 1.0) repeat in many sentences.

\paragraph{Training and testing.}
The Linear Regression is conducted on the full original data. The SVM classification is done on the sentences marked with the helpful and unhelpful classes only (neutral ignored).

\paragraph{Evaluation.}
We use the metrics reported in \citep{gamzu2021helpfulsentences} for the regression task, and accuracy for the binary classification task.

\paragraph{More results.}
\autoref{tab_classification_results_full} and \autoref{tab_helpful_sentences_results_full} extend \autoref{tab_classification_results} and \autoref{tab_helpful_sentences_results} respectively, including all \taxtype{} sets defined in \autoref{tab_taxonomy_groups}.

\subsection{Prediction of Sentiment Polarity}
\label{sec_appendix_experiments_sentiment}

\paragraph{Data used.}
As in \S{\ref{sec_appendix_experiments_review_helpfulness}}, we use the subset of products from \citet{gamzu2021helpfulsentences} for convenience. In order to prevent any bias from the helpfulness signal, we randomly sampled 5000 reviews (out of 58205) from the data without any feature pertaining to up-votes and down-votes. The review rating distribution is: $\{\texttt{5}: 3337, \texttt{4}: 675, \texttt{3}: 321, \texttt{2}: 219, \texttt{1}: 448\}$, and the polarity hence distributes to: $\{\texttt{positive}: 4012, \texttt{negative}: 988\}$. There are an average of $4.05$ sentences per review.

\paragraph{Training and testing.}
As in \S\ref{sec_appendix_experiments_review_helpfulness}, we use 50 iterations of cross validation, with a train set of 70\% and test of 30\%, averaging results over the 50 iterations. The procedure is repeated for the different \taxtype{} sets separately.

\paragraph{Evaluation.}
As in \S\ref{sec_appendix_experiments_review_helpfulness}, we report the accuracy measure.

\paragraph{More results.}
\autoref{tab_classification_results_full} extends \autoref{tab_classification_results}, including all \taxtype{} sets defined in \autoref{tab_taxonomy_groups}.

\subsection{Analysis of Reviews and Summaries}
\label{sec_appendix_experiments_analysis}

\paragraph{Data used.}
We iterated over the 31K products in the AmaSum dataset \citep{brazinskas2021amasum}, and filtered out the products without any product category assigned to it. Since products in this dataset are given several hierarchical category options, we heuristically assigned a category by manually clustering related category labels to some general ones. Then for 5 categories (``Books'', ``Electronics'', ``Apparel'', ``Toys and Games'', ``Pet Supplies'') with over 30 products, and chosen manually by their differing aspect-level characteristics, we randomly sampled 20 products. For these products we collected all their reviews and one reference summary (there are rare cases in AmaSum with more than one reference summary for a product). Overall there are 100 products with an average of $77.3$ reviews per product ($7729$ reviews total), $4.2$ sentences per review, and $7.1$ sentences per summary.

\paragraph{Analysis.}
Like in other experiments, a review/summary level vector is the average of its sentence vectors. Review/summary level vectors are then averaged to get the final two vectors to compare (in Figure \ref{fig_review_vs_summaries_vectors}).

For the rhetorical structure analysis, the 6-sentence reviews behave similarly to reviews with other lengths (Figure \ref{fig_review_structure}). Less than 6 sentences does not emphasize the behavior visually as clearly. There are $763$ (out of $7729$) reviews with 6 sentences. For the analysis on summaries, the 7-sentence summaries had the highest number of instances ($24$ out of $100$), and it is visually easier to see the patterns in the data due to the summaries containing 3 subsections (\emph{verdict}, \emph{pros}, \emph{cons}), although other length summaries behave similarly. Here, the sentence vectors at each review/summary position are averaged in order to plot the graphs. In the figures, only the \taxtypes{} with observable changes that have probabilities above 0.2 throughout the review/summary are displayed.

\section{Predictor Evaluation with Specific \taxtypes{}}
\label{sec_appendix_predictor_eval_type_specific}

In addition to the standard evaluation of our \taxtypes{} prediction model in Section \ref{sec_prediction_quality}, we additionally assessed our model's performance on benchmarks of specific \taxtypes{} already identified in previous work, namely \textit{tip}, \textit{opinion}, and \textit{opinion with reason}. The favorable results here only place further emphasis on the reliability of our model, which gives confidence to perform our analyses in Sections \ref{sec_experiments} and \ref{sec_analysis}. \textbf{This assessment is only supplemental to our evaluation in Section \ref{sec_prediction_quality}.}

We obtain existing annotated datasets and establish training sets for the sole purpose of setting a prediction threshold for one of our \taxtypes{}, which is then used as the predicted label for the annotated task.
We experiment with tuning over the entire training set as well as with much smaller subsets of 100 sentences.

\subsection{Tip Classification Evaluation}
\label{sec_appendix_experiments_tips}

Product tips are generally defined as short, concise, practical, self-contained pieces of advice on a product~\citep{hirsch2021producttips}, for example: \say{Read the card that came with it and see all the red flags!}. While previous work~\citep{hirsch2021producttips, farber2022tips} characterized tips into finer-grained sub-types, our definition assumes a generic definition.

\paragraph{Data used.}
We use the data annotated by~\citet{hirsch2021producttips} over Amazon reviews for non-tips or tips. The data used by \citet{hirsch2021producttips} for their experiments (3059 tips and 48,870 non-tip sentences) is slightly different from the data we used (3848 tips and 81,323 non-tips), since we do not apply the initial rule-based filtering that they enforce.

We noticed that some sentences were annotated in the dataset as tips, even though we do not view them as such, e.g.,~\say{The ring that holds the card together is kind of flimsy}. Since this was especially true under the \textit{warning} sub-type, we removed these sentences from the data, to better represent our notion of a \textit{tip}.

Our main goal is to show that our predictor performs decently, and not that we provide a better solution than \citet{hirsch2021producttips}. Hence, exacting the data distribution is not a major concern. Of the available data, we use all 3,848 tip-sentences and sample twice as many non-tip sentences (out of the 81,323).

The dataset includes reviews from 5 categories: ``Musical Instruments'', ``Baby'', ``Toys and Games'', ``Tools and Home Improvement'', and ``Sports and Outdoors''.

\paragraph{Training and testing.}
We follow the cross-validation \citep{xu2001crossVal} procedure of \citet{hirsch2021producttips} to train (find the best tip-\taxtype{} threshold) and to respectively evaluate tip identification using the found threshold. For 50 iterations, all tips are used (or \textit{warning} sub-types are removed) and the same amount of non-tips are sampled (out of the many available). Then the data is split to 80\%/20\% train/test splits. In case a train set size is forced, e.g., only 100 samples, then it is sampled from the training data. The train set is used to find the optimal threshold using Youden's J statistic on the ROC curve \citep{youden1950youdenj}. Then the test set is used to compute the different evaluation metrics. The results are averaged over the 50 iterations and reported, and the bootstrapping method \citep{efron1992bootstrap} is used to compute confidence intervals at alpha=0.025 (95\% confidence percentile). Since the results in \citet{hirsch2021producttips} do not report confidence intervals, we cannot fully compare to their results, however the differences are large enough to assume significant differences, as viewed in our findings.

\paragraph{Evaluation.}
\citet{hirsch2021producttips} report Recall@Precision scores, i.e., setting a specific precision value and producing the corresponding recall value. This approach is used since there is a tradeoff between presenting tips for more products and being confident about the tips presented. We also present the $F_1$ of the tip/no-tip prediction, as common for classification tasks.

\subsection{Opinions Classification Evaluation}
\label{sec_appendix_experiments_opinions}
Product reviews are rich in opinions and are often more convincing when a reason is provided for the opinion. 
A persuasive saying, whether opinionated or objective, is a type of \textit{argument}~\citep{ecklekohler2015arguments}, a class found to be helpful for predicting review helpfulness~\citep{liu2017argsForHelpfullness,passon2018helpfulness,chen2022argumentMiningForHelpfulness}. We thus turn to an argument mining dataset as an assessment benchmark.

\paragraph{Data used.}
The $AM^2$~\citep{chen2022argumentMiningForHelpfulness} dataset is annotated at the clause level for various sub-types of subjective and objective arguments and reasons.
The data contains 878 reviews (pre-split to train and test sets) from 693 products in the ``Headphones'' category, and only sentences that are argumentative are kept in each review. Each review is broken down to its clauses (a sentence or smaller). A clause is annotated as ``Policy'', ``Value'', ``Fact'' or ``Testimony'', where the first two are subjective and the latter two are objective. In addition, a clause can be marked as ``Reason'' or ``Evidence'', where the former provides support for a subjective clause, and the latter for an objective clause. The support clauses are linked to relevant clauses that they support.

We automatically parse and re-label this data and create full-sentence-level instances for \textit{opinion} and \textit{opinion with reason} \taxtypes{}. If a sentence contains a subjective clause, it is marked as an \textit{opinion}, and if it also contains a support clause then it is also marked as an \textit{opinion with reason}. Otherwise a sentence is neither.

We end up with 3132 (2133 train, 999 test) \textit{opinion} sentences, of which 363 (249 train, 114 test) are also \textit{opinion with reason}, and 1359 (972 train, 387 test) non-opinion sentences.

\paragraph{Training and testing.}
We separately classify the \textit{opinion} \taxtype{} and the \textit{opinion with reason} \taxtypes{} against the non-opinionated class. As in the case of \textit{tip}s, we use the train set to find the best threshold for the relevant \taxtype{}, and then evaluate on the test set. We also experimented with sampling just 100 train instances for tuning the threshold.

\paragraph{Evaluation.}
As common in classification tasks, we report the $F_1$ measure.

\begin{table}[t]
    \centering
    \resizebox{\columnwidth}{!}{
    \begin{tabular}{lcc}
        \toprule
        Classification Evaluation & Train Size & Test Size \\
        \midrule
        \textit{Tip} - \citet{hirsch2021producttips}          & 4894 & 1224 \\
        \textit{Tip} - Our experiments                        & 6156 & 1540 \\
        \textit{Tip} - Our experiments, no \textit{warning}s  & 4140 or 100 & 1036 \\
        \textit{Opinion} - Our experiments                    & 3105 or 100 & 1386 \\
        \textit{Opinion with Reason} - Our experiments        & 1221 or 100 & 501 \\
        \midrule
        Standard evaluation (All \taxtypes{} - \S \ref{sec_prediction_quality})                                   & 300 (dev set) & 240 \\
        \bottomrule
    \end{tabular}}
    \caption{The dataset sizes used in the classification evaluations in Appendix \ref{sec_appendix_predictor_eval_type_specific} and Section \ref{sec_prediction_quality}.}
    \label{tab_predictor_dataset_sizes}
\end{table}

\subsection{Results}
\label{sec_appendix_experiments_specific_type_results}
\autoref{tab_predictor_eval} presents the classification results on the three specific \taxtype{} evaluations.
Our zero-shot classifier, which only uses the training sets to find an optimal threshold for the corresponding \taxtype{}, is able to identify \textit{tip}s and \textit{opinion}s effectively, and \textit{opinion with reason} fairly well, on their respective benchmarks.\footnote{\citet{chen2022argumentMiningForHelpfulness} do not provide any intrinsic baseline results on opinion classification to which we can compare. In addition, we manipulated the original data from clause-level to sentence-level.} Moreover, limiting the training sets to only 100 samples appears to hardly have any effect on the quality of the model, showing its robustness to paucity of labeled data.

For the \textit{tip} task, we see in \autoref{tab_tips_results_vs_baseline_full} that the predictor performs on par or much better than existing supervised baselines from \citet{hirsch2021producttips}. Although our definition of \emph{tip} differs slightly from that used in annotating the benchmark, our predictor is still reliable and useful.
When the \textit{warning} subtype of a \textit{tip} is removed from the data (which, as mentioned before, generally do not fit our notion of a tip), the results dramatically improve.

\begin{table}[t]
    \centering
    \begin{tabular}{lcc}
        \toprule
         & \multicolumn{2}{c}{$|$Train$|$} \\
        Our Classifier on \taxtype{} & 100 & full \\
        \midrule
        \textit{Tip} (no \textit{warning}s) & 70.0 & 71.0 \\
        \textit{Opinion}                    & 85.8 & 87.6 \\
        \textit{Opinion w/Reason}           & 62.8 & 62.8 \\
        \bottomrule
    \end{tabular}
    \caption{$F_1$ on \taxtype{}-specific datasets when using 100 instances, or all instances in the training set to tune the \taxtype{} threshold.}
    \label{tab_predictor_eval}
\end{table}

\begin{table}[t]
    \centering
    \resizebox{\columnwidth}{!}{
    \begin{tabular}{lrrr}
        \toprule
        & \multicolumn{3}{c}{Recall@Precision} \\
        \textit{Tip} Classifier & 75 & 80 & 85 \\
        \midrule
        \citeauthor{hirsch2021producttips} Best Baseline & 43.4 & 29.0 & 16.8 \\
        \citeauthor{hirsch2021producttips} Best (BERT)   & 70.5 & 58.1 & 36.1 \\
        Ours                                       & 42.6 & 32.6 & 32.2 \\
        Ours* (no \textit{warning}s)               & 63.7 & 49.7 & 40.3 \\
        \bottomrule
    \end{tabular}}
    \caption{Results on the \textit{Tip} classification benchmark of \citet{hirsch2021producttips}. The * signifies we only use 100 train instances for tuning the \textit{Tip} \taxtype{} threshold. Using the full train set improves results insignificantly.}
    \label{tab_tips_results_vs_baseline_full}
\end{table}

\begin{table*}[t]
\centering
    \resizebox{\linewidth}{!}{
    \begin{tabular}{llll}
\hline
\textbf{\taxtype{}}   &	\textbf{Internal Definition}	&	\textbf{Prompt Question for LLM}	&	\textbf{Sentence Example} \\
\hline
\hline
opinion					&	\begin{tabular}[c]{@{}l@{}}a subjective expression regarding\\the product or something else\end{tabular}	&	\begin{tabular}[c]{@{}l@{}}does the sentence express\\an opinion about anything\end{tabular}	&	\begin{tabular}[c]{@{}l@{}}I love this so much that I reordered\\another one online. \end{tabular} \\
\hline
opinion\_with\_reason	&   \begin{tabular}[c]{@{}l@{}}a subjective expression regarding\\the product or something else,\\along with reasoning for the viewpoint\end{tabular}   &	\begin{tabular}[c]{@{}l@{}}does the sentence express an\\opinion about anything and also\\provide reasoning for it \end{tabular} & \begin{tabular}[c]{@{}l@{}}The book is well written in\\an easy to read format.\end{tabular} \\
\hline
improvement\_desire		&	\begin{tabular}[c]{@{}l@{}}something the customer wishes\\that the product could have had\\for improvement\end{tabular}	&	\begin{tabular}[c]{@{}l@{}}does the sentence say how\\the product could be improved\end{tabular}	&	\begin{tabular}[c]{@{}l@{}}More new tiles would have been\\nice for this new release. \end{tabular} \\
\hline
comparative				&	\begin{tabular}[c]{@{}l@{}}compares to another product,\\another version of the product,\\or family of products\end{tabular}	&	\begin{tabular}[c]{@{}l@{}}does the sentence compare\\to another product\end{tabular}	&	\begin{tabular}[c]{@{}l@{}}Hopefully Broan makes better motors. \end{tabular} \\
\hline
comparative\_general	&	\begin{tabular}[c]{@{}l@{}}something that compares the\\product generally to other things\end{tabular}	&	\begin{tabular}[c]{@{}l@{}}does the sentence describe\\something that compares the\\product generally to something\\that is not a product\end{tabular}	&	\begin{tabular}[c]{@{}l@{}}This is not a low carb or Paleo diet. \end{tabular} \\
\hline
buy\_decision			&	\begin{tabular}[c]{@{}l@{}}says something straightforward\\about acquiring or not\\acquiring the product\end{tabular}	&	\begin{tabular}[c]{@{}l@{}}does the sentence explicitly\\talk about buying the product\end{tabular}	&	\begin{tabular}[c]{@{}l@{}}I'll be buying a copy for my\\almost-2-year-old granddaughter. \end{tabular} \\
\hline
speculative				&	\begin{tabular}[c]{@{}l@{}}something the customer thinks\\will be the case with the\\product or that will happen\\because of the product\end{tabular}	&	\begin{tabular}[c]{@{}l@{}}does the sentence speculate\\about something\end{tabular}	&	\begin{tabular}[c]{@{}l@{}}I expect it to last another 2-3 years. \end{tabular} \\
\hline
personal\_usage			&	\begin{tabular}[c]{@{}l@{}}describes something that\\someone did with the product,\\sometimes describes what\\happened to the product after\\some time/use\end{tabular}	&	\begin{tabular}[c]{@{}l@{}}does the sentence describe\\how someone used the product\end{tabular}	&	\begin{tabular}[c]{@{}l@{}}I walked all over Europe in these! \end{tabular} \\
\hline
situation				&	\begin{tabular}[c]{@{}l@{}}explains a situation in which\\the product is used (usually\\relates to "personal/product usage")\end{tabular}	&	\begin{tabular}[c]{@{}l@{}}does the sentence describe\\a condition under which\\the product is used\end{tabular}	&	\begin{tabular}[c]{@{}l@{}}When a car won't jump, we can't help but\\wonder if it is the battery or these cables. \end{tabular} \\
\hline
setup					&	\begin{tabular}[c]{@{}l@{}}something about the\\setup/installation of the product\end{tabular}	&	\begin{tabular}[c]{@{}l@{}}does the sentence describe\\something about the setup\\or installation of the product\end{tabular}	&	\begin{tabular}[c]{@{}l@{}}The instructions were clear and the mounting\\went flawlessly. \end{tabular} \\
\hline
tip						&	\begin{tabular}[c]{@{}l@{}}a suggestion for what to do with\\the product\end{tabular}	&	\begin{tabular}[c]{@{}l@{}}does the sentence provide\\a tip on the product\end{tabular}	&	\begin{tabular}[c]{@{}l@{}}If I were a newby on this, I would just buy the\\older edition for a buck and use it instead. \end{tabular} \\
\hline
product\_usage			&	\begin{tabular}[c]{@{}l@{}}descibes how the product can be\\used\end{tabular}	&	\begin{tabular}[c]{@{}l@{}}does the sentence describe\\how the product can be used\end{tabular}	&	\begin{tabular}[c]{@{}l@{}}they're great for wearing at home or\\gym and doing exercises in one place \end{tabular} \\
\hline
product\_description	&	\begin{tabular}[c]{@{}l@{}}something objective about the\\product like its\\characteristics or story line\end{tabular}	&	\begin{tabular}[c]{@{}l@{}}does the sentence describe\\something objective about\\the product like its characteristics\\or its plot\end{tabular}	&	\begin{tabular}[c]{@{}l@{}}I would estimate that it weighs somewhere\\around 20-25 lbs by itself. \end{tabular} \\
\hline
price					&	\begin{tabular}[c]{@{}l@{}}talks about the price explicitly,\\possibly at a different\\retailers\end{tabular}	&	\begin{tabular}[c]{@{}l@{}}does the sentence explicitly\\talk about the price of the product\end{tabular}	&	\begin{tabular}[c]{@{}l@{}}i bought a pair from the store for 50\$\\and they are very comfortable \end{tabular} \\
\hline
compatibility			&	\begin{tabular}[c]{@{}l@{}}describes usage of the product\\along with another product\end{tabular}	&	\begin{tabular}[c]{@{}l@{}}does the sentence describe\\the compatibility of the product\\with another product\end{tabular}	&	\begin{tabular}[c]{@{}l@{}}Even though it was sold with a particular\\product as an extra battery, it\\isn't, and it doesn't fit into that product. \end{tabular} \\
\hline
personal\_info			&	\begin{tabular}[c]{@{}l@{}}something personal about\\someone, that may or may not\\have to do with the product\end{tabular}	&	\begin{tabular}[c]{@{}l@{}}does the sentence say something\\about someone\end{tabular}	&	\begin{tabular}[c]{@{}l@{}}It's just a beautiful book, one I will keep\\and pass down to my kids' children in 30 years. \end{tabular} \\
\hline
general\_info			&	\begin{tabular}[c]{@{}l@{}}general information that may\\or may not have to do with the\\product\end{tabular}	&	\begin{tabular}[c]{@{}l@{}}does the sentence describe\\general information that is not\\necesarilly in regards to the product\end{tabular}	&	\begin{tabular}[c]{@{}l@{}}There are MANY versions of ``A Christmas \\Carol'' out there. \end{tabular} \\
\hline
comparative\_seller		&	\begin{tabular}[c]{@{}l@{}}comparison between sellers of\\the product\end{tabular}	&	\begin{tabular}[c]{@{}l@{}}does the sentence compare\\between sellers of the product\end{tabular}	&	\begin{tabular}[c]{@{}l@{}}Saved a bundle buying this from Amazon\\and not the cable company. \end{tabular} \\
\hline
seller\_experience		&	\begin{tabular}[c]{@{}l@{}}something about the experience\\with the seller\end{tabular}	&	\begin{tabular}[c]{@{}l@{}}does the sentence describe\\something about the experience\\with the seller\end{tabular}	&	\begin{tabular}[c]{@{}l@{}}I contacted the seller through email\\and they were amazing. \end{tabular} \\
\hline
delivery\_experience	&	\begin{tabular}[c]{@{}l@{}}something about the experience\\of the delivery\end{tabular}	&	\begin{tabular}[c]{@{}l@{}}does the sentence describe\\the shipment of the product\end{tabular}	&	\begin{tabular}[c]{@{}l@{}}The packaging was good too. \end{tabular} \\
\hline
imagery					&	\begin{tabular}[c]{@{}l@{}}a dramatic or figurative\\description of something\end{tabular}	&	\begin{tabular}[c]{@{}l@{}}is the sentence written\\in a figurative style\end{tabular}	&	\begin{tabular}[c]{@{}l@{}}You change them out just about as frequently\\as the oil in your lawn tractor. \end{tabular} \\
\hline
sarcasm					&	\begin{tabular}[c]{@{}l@{}}the sentence has a sarcastic\\expression\end{tabular}	&	\begin{tabular}[c]{@{}l@{}}does the sentence contain\\a sarcastic expression\end{tabular}	&	\begin{tabular}[c]{@{}l@{}}Conveniently, it's not returnable either. \end{tabular} \\
\hline
rhetorical				&	\begin{tabular}[c]{@{}l@{}}something used as a filler or\\for transition, but is not\\really needed\end{tabular}	&	\begin{tabular}[c]{@{}l@{}}is the sentence rhetorical\\or used as a filler or for transition\\without any real value\end{tabular}	&	\begin{tabular}[c]{@{}l@{}}There's not much to say about products like this. \end{tabular} \\
\hline
inappropriate			&	\begin{tabular}[c]{@{}l@{}}contains content that is toxic\\or unnecessarily racy\end{tabular}	&	\begin{tabular}[c]{@{}l@{}}does the sentence contain\\content that is toxic or\\unnecessarily racy\end{tabular}	&	\begin{tabular}[c]{@{}l@{}}Buying this was a f****** waste of money. \end{tabular} \\
\hline
    \end{tabular}}
    \caption{\textbf{The full typology of review sentence \taxtypes{}}. Sentences can be labeled with several \taxtypes{}. The prompt question substring is used for filling in the prompt template (Appendix \ref{sec_appendix_predictor_model}) for tagging the corresponding \taxtype{}.}
    \label{tab_taxonomy_types}
\end{table*}

\begin{table*}[t]
    \centering
    \resizebox{\linewidth}{!}{
    \begin{tabular}{lll}
        \toprule
        \textbf{Review Sentence} & \textbf{Gold \taxtype{} Labels} & \textbf{Predicted \taxtype{} Labels} \\
        \midrule
        \midrule
        \begin{tabular}[c]{@{}l@{}}These weaves are great for\\practicing at home.\end{tabular} & \begin{tabular}[c]{@{}l@{}}opinion, product\_usage, tip\end{tabular} & \begin{tabular}[c]{@{}l@{}}\textbf{opinion}, opinion\_with\_reason,\\personal\_usage, \textbf{product\_usage},\\situation, product\_description, \textbf{tip}\end{tabular} \\
        \midrule
        \begin{tabular}[c]{@{}l@{}}I think you could get 16\\hours out of it if you\\listened on 5\% volume\\while flying in a fighter jet\\over multiple time zones\\and you calculated the\\time change.\end{tabular} & \begin{tabular}[c]{@{}l@{}}opinion, product\_description,\\sarcasm, \underline{imagery}, speculative\end{tabular} & \begin{tabular}[c]{@{}l@{}}\textbf{opinion}, opinion\_with\_reason,\\personal\_usage, product\_usage,\\\textbf{product\_description}, \textbf{speculative},\\comparative\_general, \textbf{sarcasm}\end{tabular} \\
        \midrule
        \begin{tabular}[c]{@{}l@{}}Our dog seems to like them\\just fine but they dont last\\very long because they are\\MAYBE 1/2 of what they\\advertise.\end{tabular} & \begin{tabular}[c]{@{}l@{}}opinion, product\_description,\\\underline{personal\_info}\end{tabular} & \begin{tabular}[c]{@{}l@{}}\textbf{opinion}, opinion\_with\_reason,\\personal\_usage, product\_usage,\\\textbf{product\_description}, speculative,\\price\end{tabular} \\
        \midrule
        \begin{tabular}[c]{@{}l@{}}They are plenty loud for my\\trail riding/work around\\the property!\end{tabular} & \begin{tabular}[c]{@{}l@{}}opinion, personal\_usage,\\\underline{personal\_info}, product\_description,\\situation\end{tabular} & \begin{tabular}[c]{@{}l@{}}\textbf{opinion}, opinion\_with\_reason,\\\textbf{personal\_usage}, product\_usage,\\\textbf{situation}, \textbf{product\_description}\end{tabular} \\
        \midrule
        \begin{tabular}[c]{@{}l@{}}Really, they are pretty basic\\recipes that you could find\\anywhere.\end{tabular} & \begin{tabular}[c]{@{}l@{}}opinion, speculative,\\product\_description,\\comparative\_general\end{tabular} & \begin{tabular}[c]{@{}l@{}}\textbf{opinion}, \textbf{product\_description},\\\textbf{speculative}, \textbf{comparative\_general},\\sarcasm\end{tabular} \\
        \midrule
        \begin{tabular}[c]{@{}l@{}}I bought this Hard Drive for my\\Xbox One because 500 gb\\internal was not enough seeing\\as how an average game is\\almost 50 GB alone.\end{tabular} & \begin{tabular}[c]{@{}l@{}}buy\_decision, compatibility,\\personal\_usage, \underline{general\_info},\\\underline{speculative}\end{tabular} & \begin{tabular}[c]{@{}l@{}}opinion, opinion\_with\_reason,\\\textbf{personal\_usage}, product\_usage,\\product\_description, tip,\\\textbf{compatibility}, \textbf{buy\_decisio}n\end{tabular} \\
        \midrule
        \begin{tabular}[c]{@{}l@{}}Live guide has a learning curve,\\but worth it for sure.\end{tabular} & \begin{tabular}[c]{@{}l@{}}opinion, tip, product\_usage\end{tabular} & \begin{tabular}[c]{@{}l@{}}\textbf{opinion}, opinion\_with\_reason,\\setup, \textbf{product\_usage},\\product\_description, \textbf{tip}\end{tabular} \\
        \midrule
        \begin{tabular}[c]{@{}l@{}}I love the set however it arrived\\with one of the cups that hold\\the poles completely stripped\\and I will have to either find a\\larger screw or super glue it.\end{tabular} & \begin{tabular}[c]{@{}l@{}}opinion, delivery\_experience,\\\underline{situation}, \underline{personal\_usage},\\\underline{personal\_info}, \underline{product\_description}\end{tabular} & \begin{tabular}[c]{@{}l@{}}\textbf{opinion}, opinion\_with\_reason,\\buy\_decision, \textbf{delivery\_experience}\end{tabular} \\
        \midrule
        \begin{tabular}[c]{@{}l@{}}I think I can get a dozen\\washes out of this order,\\so it's a very good value.\end{tabular} & \begin{tabular}[c]{@{}l@{}}opinion, opinion\_with\_reason,\\\underline{product\_description}, \underline{product\_usage},\\speculative\end{tabular} & \begin{tabular}[c]{@{}l@{}}\textbf{opinion}, \textbf{opinion\_with\_reason},\\personal\_info, \textbf{speculative},\\price, buy\_decision\end{tabular} \\
        \bottomrule
    \end{tabular}}
    \caption{Examples of review sentences with their gold \taxtype{} labels and predicted labels (by our predictor outlined in \S{\ref{sec_prediction_model}}). \textbf{Bold} \taxtypes{} are true-positive predictions (correctly predicted) and \underline{underlined} ones are false-negatives (wrongly missed by the predictor). In many cases, false-positive \taxtypes{} (wrongly predicted) are not irrational. The predictor is sufficiently reliable for conducting our experiments and analyses, which is our main objective.}
    \label{tab_example_annotations}
\end{table*}

\begin{table*}[t]
    \centering
    \begin{tabular}{lccc|rr|r}
        \toprule
        & & & & \multicolumn{3}{c}{\# (\%) Sentences w/ \taxtype{}} \\
        \taxtype{} & $F_1$ & Recall & Precision & Gold Test & Pred Test & Gold Val \\
        \midrule
        opinion               & 88.4 & 90.3  & 86.6 & 166 (69.0) & 174 (72.4) & 211 (70.3) \\
        opinion\_with\_reason & 46.7 & 84.8  & 32.2 & 46 (19.2)  & 121 (50.6) & N/A \\
        improvement\_desire   & 50.0 & 77.8  & 36.8 & 9 (3.8)    & 19 (7.9)   & 7 (2.3) \\
        comparative           & 68.9 & 77.8  & 61.8 & 27 (11.3)  & 35 (14.6)  & 16 (5.3) \\
        comparative\_general  & 38.2 & 65.0  & 27.1 & 20 (8.4)   & 49 (20.5)  & 16 (5.3) \\
        buy\_decision         & 69.6 & 94.1  & 55.2 & 34 (14.2)  & 58 (24.3)  & 9 (3.0) \\
        speculative           & 44.2 & 60.7  & 34.7 & 28 (11.7)  & 50 (20.9)  & 18 (6.0) \\
        personal\_usage       & 66.7 & 87.8  & 53.8 & 49 (20.5)  & 80 (33.5)  & 39 (13.0) \\
        situation             & 51.1 & 55.8  & 47.1 & 43 (18.0)  & 51 (21.3)  & 13 (4.3) \\
        setup                 & 64.9 & 85.7  & 52.2 & 14 (5.9)   & 23 (9.6)   & 5 (1.7) \\
        tip                   & 50.0 & 74.1  & 37.7 & 27 (11.3)  & 53 (22.2)  & 19 (6.3) \\
        product\_usage        & 46.2 & 75.0  & 33.3 & 44 (18.4)  & 100 (41.8) & 29 (9.7) \\
        product\_description  & 63.2 & 75.0  & 54.5 & 88 (36.8)  & 122 (51.0) & 97 (32.3) \\
        price                 & 78.4 & 90.9  & 69.0 & 22 (9.2)   & 29 (12.1)  & 7 (2.3) \\
        compatibility         & 55.3 & 86.7  & 40.6 & 15 (6.3)   & 32 (13.4)  & 10 (3.3) \\
        personal\_info        & 65.7 & 60.7  & 71.4 & 108 (45.2) & 91 (38.1)  & 59 (19.7) \\
        general\_info         & 32.3 & 62.5  & 21.7 & 9 (3.8)    & 23 (9.6)   & 29 (9.7) \\
        comparative\_seller   & 54.5 & 85.7  & 40.0 & 7 (2.9)    & 15 (6.3)   & 3 (1.0) \\
        seller\_experience    & 52.9 & 100.0 & 36.0 & 9 (3.8)    & 25 (10.5)  & 7 (2.3) \\
        delivery\_experience  & 72.7 & 80.0  & 66.7 & 15 (6.3)   & 18 (7.5)   & 4 (1.3) \\
        imagery               & 62.5 & 71.4  & 55.6 & 21 (8.8)   & 27 (11.3)  & 28 (9.3) \\
        sarcasm               & 27.6 & 44.4  & 20.0 & 9 (3.8)    & 20 (8.4)   & 7 (2.3) \\
        rhetorical            & 44.4 & 34.8  & 61.5 & 23 (9.6)   & 13 (5.4)   & 17 (5.7) \\
        inappropriate         & 66.7 & 100.0 & 50.0 & 5 (2.1)    & 10 (4.2)   & 0 (0.0) \\
        \midrule
        ALL (Avg.)            & 56.7 & 75.9  & 47.7 & \multicolumn{2}{c|}{240}                  & \multicolumn{1}{c}{300} \\
        \bottomrule
    \end{tabular}
    \caption{$F_1$ scores on the test set per \taxtype{} and overall (macro-$F_1$), as described in the evaluation procedure in \S{\ref{sec_prediction_quality}}, on our \taxtypes{} prediction model. There are 240 annotated sentences in the test set. The threshold of each \taxtype{}, applied on the score from the predictor, is decided by tuning on the development set. The count and percentage of sentences with each one of the \taxtypes{} is shown here as well, to showcase the general distribution of sentences with each \taxtype{}. The rightmost column shows the distribution of sentences in the validation set (300 sentences total). The gold and validation percentages correlate at $\rho=0.91$ (Pearson).}
    \label{tab_test_scores_per_type}
\end{table*}

\end{document}